\documentclass{article}

\usepackage{arxiv}
\usepackage{algorithm}
\usepackage{algorithmic}
\usepackage{amsmath}
\usepackage{amssymb}
\usepackage{mathtools}
\usepackage{amsthm}
\usepackage{makecell}
\usepackage[dvipsnames,svgnames]{xcolor}
\usepackage[utf8]{inputenc} 
\usepackage[T1]{fontenc}    
\usepackage{hyperref}       
\usepackage{url}            
\usepackage{booktabs}       
\usepackage{amsfonts}       
\usepackage{nicefrac}       
\usepackage{microtype}      
\usepackage{cleveref}       
\usepackage{lipsum}         
\usepackage{graphicx}
\usepackage{natbib}
\usepackage{doi}
\usepackage{bm}
\usepackage{enumitem}
\usepackage{mathrsfs}
\usepackage{wrapfig}
\usepackage{multirow}

\title{Self-Discriminative Modeling for \\Anomalous Graph Detection}

\author{Jinyu~Cai$^\dagger$\\
	Institute of Data Science,\\
        National University of Singapore\\
	\texttt{jinyucai1995@gmail.com} \\
	\And
	Yunhe~Zhang$^\dagger$\\
        Shenzhen Research Institute of Big Data\\
        Shenzhen, China, 518172 \\
	\texttt{zhangyhannie@gmail.com} \\
         \And
        Jicong Fan\thanks{Corresponding author. $^\dagger$Equal contribution.} \\
	School of Data Science\\
	The Chinese University of Hong Kong, Shenzhen\\
        Shenzhen Research Institute of Big Data\\
        Shenzhen, China, 518172 \\
	\texttt{fanjicong@cuhk.edu.cn} \\
}

\renewcommand{\shorttitle}{Self-Discriminative Modeling for Anomalous Graph Detection}

\hypersetup{
pdftitle={Self-Discriminative Modeling for Anomalous Graph Detection},
pdfsubject={Graph-level Anomaly Detection},
pdfauthor={Jinyu Cai, Yunhe Zhang, Jicong Fan},
pdfkeywords={Anomaly Detection, Graph-level Learning, Graph Neural Network},
}

\begin{document}
\maketitle

\begin{abstract}
This paper studies the problem of detecting anomalous graphs using a machine learning model trained on only normal graphs, which has many applications in molecule, biology, and social network data analysis. We present a self-discriminative modeling framework for anomalous graph detection. The key idea, mathematically and numerically illustrated, is to learn a discriminator (classifier) from the given normal graphs together with pseudo-anomalous graphs generated by a model jointly trained, where we never use any true anomalous graphs and we hope that the generated pseudo-anomalous graphs interpolate between normal ones and (real) anomalous ones. Under the framework, we provide three algorithms with different computational efficiencies and stabilities for anomalous graph detection. The three algorithms are compared with several state-of-the-art graph-level anomaly detection baselines on nine popular graph datasets (four with small size and five with moderate size) and show significant improvement in terms of AUC. The success of our algorithms stems from the integration of the discriminative classifier and the well-posed pseudo-anomalous graphs, which provide new insights for anomaly detection.  Moreover, we investigate our algorithms for large-scale imbalanced graph datasets. Surprisingly, our algorithms, though fully unsupervised, are able to significantly outperform supervised learning algorithms of anomalous graph detection. The corresponding reason is also analyzed.
\end{abstract}


\section{Introduction}
\label{sec1}
Graphs are widely utilized to represent complex relationships or interactions between entities in a variety of real-world contexts, such as the molecule, biology, and social networks data analysis~\citep{mislove2007measurement, li2021graph}. By capturing the topology, structure, and dynamics of underlying systems, graphs offer rich information. Machine learning-based anomaly detection~\citep{ruff2018deep,zong2018deep,ruff2021unifying,han2022adbench} is an essential research problem in the analysis of graphs~\citep{akoglu2015graph}, which can unveil intricate relationships and patterns in graph data, thereby leading to practical applications in fields like fraud detection~\citep{beutel2015graph}, network intrusion detection~\citep{chou2021survey}, and molecules identification in biological networks~\citep{ghavami2020anomaly}. Given a set of graph data, graph anomaly detection ~\citep{ma2021comprehensive} aims to identify unusual substructures or graphs in a given dataset, which exhibit abnormal patterns, structures, or behaviors compared to the majority of the graph data.

Generally, graph anomaly detection can be performed at different levels, with regards to node-level, edge-level, and graph-level respectively. Node-level and egde-level anomaly detection (AD) focus on identifying anomalous nodes or edges in a single graph~\citep{ma2021comprehensive}, which has been studied more intensively in recent years with the emergence of graph neural networks~\citep{maron2019provably,de2020natural}.
For example, \citet{ding2019deep} proposed to utilize graph convolutional networks (GCN) to learn node embeddings and then perform anomaly detection by reconstructing the node embeddings with an auto-encoder and measuring the reconstruction error. \citet{zheng2021generative} proposed a self-supervised learning based node anomaly detection method, which leverages contrastive learning and attribute reconstruction to exploit contextual information of target nodes from different views for detecting anomalies. \citet{duan2020aane} studied the anomaly detection at edge-level, and designed the anomaly-aware and adjusted-fitting loss to iteratively select and update the anomalous edges.

Different from node-level or edge-level AD that focuses on identifying anomalous nodes or edges within a single graph, graph-level anomaly detection (Graph-level AD) ~\citep{zhang2022dual,ma2022deep,qiu22raising} refers to the task of identifying abnormal graphs or subgraphs in a given dataset, which operates at the entire graph. Although there has been significant research on node-level and edge-level AD, graph-level AD has been less studied as it is a more challenging task compared to them. The reasons are as follows.
\begin{itemize}[leftmargin=*]
    \item[1)] Node-level or edge-level anomalies can be detected by analyzing the properties of individual nodes or unusual relationships between nodes in a single graph, while graph-level anomalies involve analyzing the overall structure of a graph and the composition of all nodes, which is more complex.
    \item[2)] In graph-level AD, it is difficult to clearly define what graphs are anomalous.
    In contrast, in node-level or edge-level AD, anomalous nodes or edges can often be well-defined based on simple statistical criteria, e.g., nodes or edges whose attributes have unusual values.
\end{itemize}

Nevertheless, graph-level AD is an important and useful problem that has attracted increasing research interest. \citet{zhao2021using} investigated the graph-level AD problem and proposed one-class graph isomorphism network (OCGIN). OCGIN combines the deep one-class classification (DSVDD)~\citep{ruff2018deep} with graph isomorphism network (GIN)~\citep{xupowerful}. They also explored the feasibility of graph embeddings~\citep{narayanan2017graph2vec, grohe2020word2vec} and graph kernels~\citep{shervashidze2011weisfeiler, borgwardt2005shortest} in two-stage graph-level AD. \citet{qiu22raising} proposed one-class graph transformation learning (OCGTL) to address the performance flip issue in OCGIN with neural transformation learning~\citep{qiu2021neural}, achieving a significant improvement in performance. ~\citet{ma2022deep} proposed global and local knowledge distillation (GLocalKD) for graph-level AD, which learns the normal patterns from both global and local perspectives by randomly distilling the graph and node representations. \citet{zhang2022dual} proposed imbalanced graph-level anomaly detection (iGAD) that learns a classifier to distinguish anomalies from normal graphs via graph convolution-based attribute anomaly aware network and a deep random walk kernel-based anomaly sub-structure anomaly aware network.

Despite the recent advances in graph-level AD, there are still some limitations that need to be addressed. For example, OCGIN \citep{zhao2021using} and OCGTL \citep{qiu22raising} rely on a strong assumption about the shape of the embedding distribution of graphs, i.e., assuming it to be a hypersphere, which may not always hold or be achieved in real-world scenarios. Additionally, GLocalKD \citep{ma2022deep}, OCGIN, and OCGTL require a specific definition of anomaly scores, which can be challenging to define in practice as the criteria for measuring anomalous graphs are not easy to determine. Although iGAD \citep{zhang2022dual} is a promising approach that trains a classifier to distinguish anomalies, it is a supervised approach that requires labeled data, which is often costly to obtain in some scenarios.

\begin{figure}[h!]
    \centering
    \includegraphics[height=!,width=0.9\linewidth]{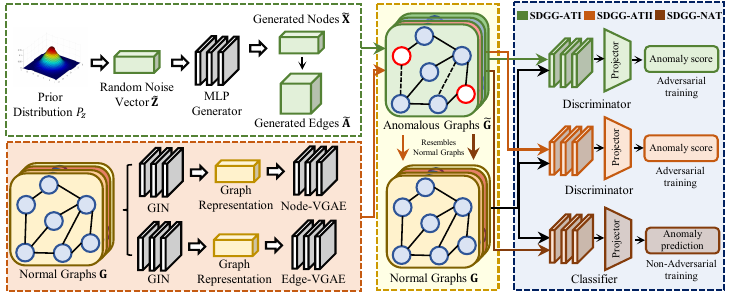}
    \vspace{-5pt}
    \caption{Network structure of the propose SDGG-ATI, SDGG-ATII, and SDGG-NAT.}
    \label{fig:network}
    \vspace{-1pt}
\end{figure}

In this paper, we propose a novel framework for graph-level anomaly detection. The key idea is distinguishing normal graphs from the generated pseudo-anomalous graphs that interpolate between normal ones and (real) anomalous ones. To generate such pseudo-anomalous graphs, we introduce two approaches: 1) training a generator using random noise from a latent distribution, and 2) training a perturbator to create anomalies from normal graphs. Both approaches leverage adversarial training, incorporating a discriminator to differentiate between normal samples and pseudo-anomalous samples. Moreover, we propose a non-adversarial approach to enhance model stability as well as accuracy. Based on each of the three approaches, importantly, the classifier serves as the anomaly detector and adaptively learns the decision boundary between normality and abnormality. Figure~\ref{fig:network} presents the network structure of the proposed methods. Our contributions are:
\vspace{-5pt}
\begin{itemize}[leftmargin=*]
    \item We propose a novel and efficient graph-level anomaly detection framework that revolves around training a discriminator (classifier) to effectively distinguish normal graphs from well-posed pseudo-anomalous graphs.
    \item We introduce two adversarial approaches to produce pseudo-anomalous graphs that closely resemble normal graphs but are more similar to anomalous graphs, where a discriminator is learned jointly.
    \item We introduce a non-adversarial approach that learns a classifier to distinguish between normal graphs and pseudo-anomalous (adaptively perturbed normal graphs). Compared to the previous two approaches, this one has higher model training stability and anomaly detection accuracy.
\end{itemize}
Our algorithms are compared with the state-of-the-art methods of graph-level AD on 13 benchmark graph datasets and show significant improvement. Particularly, on large-scale imbalanced graph datasets, our algorithms, though fully unsupervised, outperform many supervised AD algorithms.

\section{Self-Discriminative Modeling for Anomalous Graph Detection}
\label{sec2}

\subsection{Problem Formulation and Motivation} Let $\mathbb{G} = \{G_{1}, \dots, G_{N}\}$ be a graph dataset comprising $N$ graphs, where a single graph $G_{i} = \{V_{i}, E_{i}\}$ contains a node set $V_{i}$ and an edge set $E_{i}$. The adjacency matrix of $G_i$ is denoted by $\mathbf{A}_{i} \in \{0,1\}^{n_i\times n_i}$, where $n_i=\vert V_i\vert$. The feature matrix of nodes of $G_i$ is denoted by $\mathbf{X}_{i}\in \mathbb{R}^{n_i\times d}$. Suppose the graphs in $\mathbb{G}$ are normal graphs, we want to learn a model from $\mathbb{G}$ to determine whether a new graph $G_{\text{new}}$ is normal or abnormal. This problem is called \textit{anomalous graph detection} (AGD)\footnote{Note that this is an unsupervised learning problem, of which the training data do not contain any anomalous graphs. There are also supervised and semi-supervised settings~\citep{ruff20deep, zhang2022dual}.}. A fundamental assumption of the AGD problem is that $G_i,\ldots, G_N$ are drawn from some unknown distribution $\mathscr{D}$ (deemed as a normal distribution) while any graphs drawn from any other distributions (denoted as $\tilde{\mathscr{D}}$) are anomalous, where there is no overlap between $\mathscr{D}$ and all possible $\tilde{\mathscr{D}}$.

\begin{wrapfigure}{r}{0.5\textwidth}
    \centering
\includegraphics[width=1\linewidth,height=!]{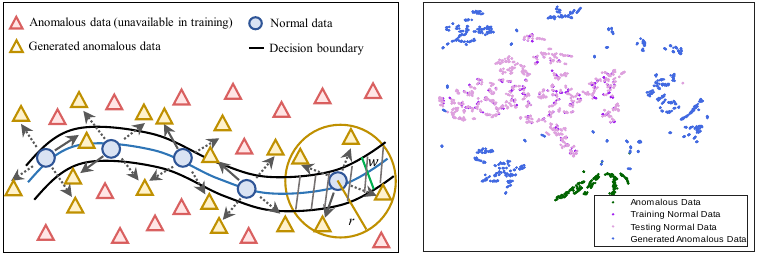}\\
    \caption{Motivation of our method (left: toy example; right: real data from AIDS (see Table \ref{Dataset}).}
    \label{fig:idea}
\vspace{-5pt}
\end{wrapfigure}

The AGD problem can be regarded as a binary classification problem, i.e., justifying $G\sim{\mathscr{D}}$ or $G\sim\tilde{\mathscr{D}}$. We want to learn a classifier $f$ from only $\mathbb{G}$ to distinguish between $G$ drawn from $\mathscr{D}$ and $\tilde{G}$ drawn from $\tilde{\mathscr{D}}$. The difficulty is that $\tilde{\mathscr{D}}$ is totally unknown. Then we need estimate $\tilde{\mathscr{D}}$ from $\mathbb{G}$ or at least generate some samples drawn from a subset of $\tilde{\mathscr{D}}$ using $\mathbb{G}$. We may solve the following problem
\begin{equation}\label{eq_theta_G_0}
\begin{aligned}
\underset{\theta,\tilde{\mathscr{D}}_s}{\textrm{minimize}}~\mathop{\mathbb{E}}_{G\sim \mathscr{D}}\ell(y,{f_{\theta}(G)})+\mathop{\mathbb{E}}_{\tilde{G}\sim \tilde{\mathscr{D}}_s\subseteq\tilde{\mathscr{D}}}\ell(\tilde{y},{f_{\theta}(\tilde{G})}),\quad
\text{subject to}~~~\text{dist}(\mathscr{D},\tilde{\mathscr{D}}_s)\leq \epsilon,
\end{aligned}
\end{equation}
where $y \equiv 0$ and $\tilde{y}\equiv 1$ denote the labels of normal and anomalous graphs respectively, $f_{\theta}(\cdot)$ denotes a classifier (e.g. a neural network) parameterized with $\theta$, and $\ell(\cdot)$ denotes the loss function.
The constraint in \eqref{eq_theta_G_0} means that $\mathscr{D}$ and $\tilde{\mathscr{D}}_{s}$ should be close enough with respect to a distance metric $\text{dist}(\cdot,\cdot)$, where $\epsilon>0$ is a small constant. However, in \eqref{eq_theta_G_0}, $\tilde{\mathscr{D}}$ is still unknown and the condition $\tilde{\mathscr{D}}_s$ does not overlap with $\mathscr{D}$ is too strong. Even when $\tilde{\mathscr{D}}_s$ and $\mathscr{D}$ overlap with each other, the learned $f_{\theta}$ could be still effective, provided that the decision boundary encloses $\mathscr{D}$ compactly (to be shown in Figure \ref{fig:idea}). Therefore, instead of \eqref{eq_theta_G_0}, we propose to solve
\begin{equation}\label{eq_theta_G}
\begin{aligned}
\underset{\theta,\phi}{\textrm{minimize}}~\mathop{\mathbb{E}}_{G\sim \mathscr{D}}\ell(y,{f_{\theta}(G)})+\mathop{\mathbb{E}}_{\tilde{G}\sim g_{\phi}(G), G\sim\mathscr{D}}\ell(\tilde{y},{f_{\theta}(\tilde{G})}),
\end{aligned}
\end{equation}
where $g_{\phi}$ converts a normal graph to a distribution of pseudo-anomalous graphs.

As shown in Figure~\ref{fig:idea},  the first plot summarizes the motivation of \eqref{eq_theta_G}: we hope that the generated pseudo-anomalous graphs interpolate between normal ones and (real) anomalous ones. Specifically, in the first plot, blue points represent normal training data, roughly lying on a (blue) curve. $g_\phi$ perturbs each normal graph randomly to generate one or more pseudo-anomalous graphs. We see that most pseudo-anomalous graphs are far from the blue curve, which can be theoretically proved as follows.
Let's consider a more general case in $d$-dimension space. The volume of the shadowed region (between the two black curves in 2D) in the radius-$r$ hypersphere (the yellow circle in 2D) can be approximated by $\prod_{i=1}^dw_i$, where $w_1=\cdots=w_\alpha=2r$ and $1\leq\alpha<d$.
Then the ratio of expected numbers of pseudo-anomalous graphs in the shadowed region and the unshadowed region in the hypersphere is computed as
\begin{equation}
 \eta =  \frac{(2r)^{\alpha}\prod_{i=\alpha+1}^{d}{w}_{i}}{\frac{\pi^{{d}/{2}}r^{d}}{\Gamma(1+{d}/{2})}-(2r)^{\alpha}\prod_{i=\alpha+1}^{d}{w}_{i}},
\end{equation}
where we have, WLOG, assumed that the points distribute uniformly. Particularly, when $d=2, \alpha =1$, we have $\eta = \frac{2w}{\pi r-2w}$.
We see that $\eta$ decreases when $w_i$ decreases or $r$ increases, where $r$ is related to the variation of pseudo anomalous graphs. We can conclude that most pseudo-anomalous graphs are outside the shadowed region when there are some small $w_i$, namely, the latent dimension of the normal data is much lower than the ambient dimension. Therefore, a classifier that can distinguish between the normal training data and most of the pseudo-anomalous graphs is sufficient to be a detector for anomalous graphs. The second plot in Figure \ref{fig:idea} is the t-SNE visualization of our method on a real dataset and highlights the successful learning of a useful decision boundary: the generated anomalous graphs are surrounding the normal ones, alongside the (real) anomalous ones. More real examples are in Figure \ref{fig:emb_score_visualization} and the supplement.
We call \eqref{eq_theta_G} \textbf{Self-Discriminative Graph Generation} (SDGG) based AD. In the following three sections, we will show how to approximately solve \eqref{eq_theta_G}.

\subsection{Self-Discriminative Modeling: SDGG-ATI}
We first present a GAN-based approach to generate pseudo-anomalous graphs. The model consists of a graph generator $\mathcal{G}_{\phi}$ and a graph discriminator $\mathcal{D}_{\omega}$, which are alternatively trained in an adversarial manner. The generator tries to produce fake (pseudo-anomalous) graphs (containing nodes and edges generation) that can fool the discriminator, while the discriminator tries to differentiate between anomalous and normal graphs. Specifically, the generator $\mathcal{G}_{\phi}$ generates nodes and edges to form a fake graph set $\tilde{\mathbb{G}} = \{\tilde{G}_{1}, \dots, \tilde{G}_{N}\}$. We first sample random variable $\tilde{\mathbf{Z}}$ from a latent distribution $\mathbb{P}_{\tilde{\mathbf{Z}}} :=\mathcal{N}(\bm{0}, \bm{1})$ and construct the adjacency matrix as follows
\begin{eqnarray}
\tilde{\mathbf{A}} = \mathcal{T}(\tilde{\mathbf{X}}\tilde{\mathbf{X}}^{\top}),\quad
\tilde{\mathbf{X}} =  \mathcal{G}_{\phi}(\tilde{\mathbf{Z}}),\quad
\tilde{\mathbf{Z}}\sim \mathbb{P}_{\tilde{\mathbf{Z}}},
\label{sample-latent}
\end{eqnarray}
where $\mathcal{G}_{\phi}$ is an MLP-based generator that maps the random latent variable $\tilde{\mathbf{Z}}\in \mathbb{R}^{N\times d}$ to the anomalous node attributes, and $\mathcal{T}:\mathbb{R}\rightarrow [0,1]$ denotes an element-wise transformation function, e.g. $\text{Sigmoid}(\cdot)$. In this way, we generate an anomalous graph set $\tilde{\mathbb{G}}$ with the generator $\mathcal{G}_{\phi}$. We then introduce a discriminator $\mathcal{D}_{\omega}$, which takes the anomalous graphs $\tilde{\mathbb{G}}$ and normal graphs $\mathbb{G}$ as input, and aims to effectively distinguish between them. To fully exploit the structural information of graphs,
$\mathcal{D}_{\omega}$ is expected to be a GNN-based network. Specifically, we leverage GIN~\citep{xupowerful} as the backbone network of the discriminator $\mathcal{D}_{\omega}$ to learn ideal graph-level representations for graph data. Assume we have an input graph $G_{i}$, the latent features $\mathbf{h}^{(k)}(v)$ of node $v$ in the $k$-th layer of GIN can be obtained by aggregating the learned features from its neighboring nodes in the $(k-1)$-th layer, which can be formulated as
\begin{eqnarray}
\mathbf{h}^{(k)}(v) = \delta(\mathrm{COMBINE}(\mathbf{h}^{(k-1)}(v),\mathrm{AGGREGATE}(\{\mathbf{h}^{(k-1)}(u),u \in\mathcal{C}(v)\}))),
\label{k-layer_feature}
\end{eqnarray}
where $\mathcal{C}(v)$ denotes the neighbor set of node $v$, and $\delta(\cdot)$ is a non-linear activation function such as ReLU. $\mathrm{AGGREGATE}(\cdot)$ function combines the features of neighboring nodes in $\mathcal{C}(v)$, and $\mathrm{COMBINE}(\cdot)$ function combines the features from the previous layer and the aggregated neighborhood information to obtain the current layer's features. Note that the attribute $\mathbf{x}_{v}$ of node $v$ serves as the initial features, i.e., $\mathbf{h}^{(0)}(v) = \mathbf{x}_{v}$. Then the graph-level representation of graph $G_{i}$ can be derived as follows:
\begin{eqnarray}
\mathbf{h}_{G_{i}} = \mathcal{R}(\{\mathrm{CONCAT}(\mathbf{h}^{(k)}(v),k\in \{1, \dots, K\})\},v \in G_{i}),
\label{graph-level-feature}
\end{eqnarray}
where $\mathrm{CONCAT}(\cdot)$ function concatenates the representations learned in each GIN layer, and $\mathcal{R}(\cdot)$ denotes the max-readout function that aggregates the node features into a graph-level representation. Consequently, we can learn the graph-level representations $\mathbf{H}_{\mathbb{G}}$ and $\mathbf{H}_{\tilde{\mathbb{G}}}$ for normal and pseudo-anomalous graphs, and train the discriminator to distinguish them as much as possible. The generator $\mathcal{G}_{\phi}$ and discriminator $\mathcal{D}_{\omega}$ are alternatively optimized with a min-max game as follows:
\begin{eqnarray}
\underset{\phi}{\min}~\underset{\omega}{\max} \underset{\mathbf{X}_{i}, \mathbf{A}_{i}\sim \mathbb{P}_{\mathbb{G}}}{\mathbb{E}} [\mathcal{D}_{\omega}(\mathbf{X}_{i}, \mathbf{A}_{i})]- \underset{\tilde{\mathbf{Z}}_{i}\sim \mathbb{P}_{\tilde{\mathbf{Z}}}}{\mathbb{E}}[\mathcal{D}_{\omega}(\mathcal{G}_{\phi}(\tilde{\mathbf{Z}}_{i}), \mathcal{T}(\mathcal{G}_{\phi}(\tilde{\mathbf{Z}}_{i}) \mathcal{G}_{\phi}(\tilde{\mathbf{Z}}_{i})^{\top}))],
\label{gan-function}
\end{eqnarray}
where $\mathbb{P}_{\mathbb{G}}$ denotes the normal graph data distribution, and $\tilde{\mathbf{Z}}_{i}\in \mathbb{R}^{n\times d'}$ is sampled from the prior distribution $\mathbb{P}_{\tilde{\mathbf{Z}}}\sim \mathcal{N}(\bm{0},\bm{1})$.  The trained discriminator can then serve as an anomaly detector. Comparing to \eqref{eq_theta_G_0}, we see that the constraint $\text{dist}(\mathscr{D},\tilde{\mathscr{D}}_s)\leq \epsilon$ is guaranteed if $\mathcal{G}_{\phi}$ given by \eqref{gan-function} are strong enough. It is difficult to guarantee for \eqref{eq_theta_G_0} that $\tilde{\mathscr{D}}_s$ does not overlap with ${\mathscr{D}}$, which however is not compulsory because it is still possible to learn a discriminator from overlapping ${\mathscr{D}},\tilde{\mathscr{D}}_s$ to distinguish between ${\mathscr{D}}$ and $\tilde{\mathscr{D}}$. We call this method SDGG-ATI, where AT represents adversarial training. Although promising for anomalous graph detection, SDGG-ATI has the following issues.
\begin{itemize}[leftmargin=*]
\item An MLP-based generator may not effectively capture the structural information of graphs, which could impede the generation of high-quality anomalous graphs for training.
\item The interpretability of the GAN-based method is limited, as generating anomalous graphs from random noise does not necessarily ensure the generation of high-quality anomalous graphs.
\item The optimization of the GAN-based method involves a min-max game, which can lead to instability during training. Besides, the competition between the generator and discriminator may result in mode collapse, leading to the generation of poor-quality anomalous graphs.
\end{itemize}

\subsection{Self-Discriminative Modeling: SDGG-ATII}
To address the first two issues of SDGG-ATI, we propose a variant of our SDGG-ATI, which can leverage the structural information, and further provide more explicit guidance for the generator $\mathcal{G}_{\phi}$, ensuring the generation of high-quality anomalous graphs that closely resemble normal ones but can still be distinguished by the discriminator.  Specifically, we use the GIN-based VGAE network as the generator $\mathcal{G}_{\phi}$, which consists of a Node-VGAE and an Edge-VGAE~\citep{kipf2016variational}, to learn anomalous graphs. The Node-VGAE aims to generate anomalous attributes $\tilde{\mathbf{X}}$, while the Edge-VGAE which does not include a decoder, aims to generate adjacency matrix $\tilde{\mathbf{A}}$. Instead of sampling the input of $\mathcal{G}_{\phi}$ from the latent distribution $\mathbb{P}_{\tilde{\mathbf{Z}}}$, we take the normal graph set $\mathbb{G}$ as the input of $\mathcal{G}_{\phi}$, with the aim of generating anomalous graphs $\tilde{\mathbb{G}}$ that are close to $\mathbb{G}$ but are expressive pseudo-anomalous graphs.

Here we only describe Node-VGAE, as it differs from Edge-VGAE just in the existence of a decoder. We first learns the graph-level representation $\mathbf{H}_{\mathbb{G}}$ for the input graphs $\mathbb{G} = \{G_{1}, \dots, G_{N}\}$ by Eq.~\eqref{k-layer_feature} and \eqref{graph-level-feature}, where $G_{i} = \{\mathbf{X}_{i}, \mathbf{A}_{i}\}$. Next, we map the graph-level representation into a latent Gaussian distribution $\mathcal{N}(\bm{\mu},\bm{\sigma}^{2})$ as in VGAE, where the means $\bm{\mu}$ and deviations $\bm{\sigma}$ are defined as follows:
\begin{eqnarray}
\bm{\mu}= \text{GIN}_{\bm{\mu}}(\mathbf{H}_{\mathbb{G}},\mathbf{A}), \ \bm{\sigma}= \exp(\text{GIN}_{\bm{\sigma}}(\mathbf{H}_{\mathbb{G}},\mathbf{A})),
\label{mu-sigma}
\end{eqnarray}
 where $\bm{\mu}$ and $\bm{\sigma}$ can explicitly define an inference model that we can sample latent graph representations $\mathbf{Z}_{\mathbb{G}}$ from it as follows:
\begin{eqnarray}
q(\mathbf{Z}_{\mathbb{G}}|\mathbf{H}_{\mathbb{G}},\mathbf{A}) = \prod_{i=1}^{N} q(\mathbf{Z}_{G_{i}}|\mathbf{H}_{\mathbb{G}},\mathbf{A}), \ \ q(\mathbf{Z}_{G_{i}}|\mathbf{H}_{\mathbb{G}},\mathbf{A}) =\mathcal{N}(\mathbf{Z}_{G_{i}}|\bm{\mu}_{i}, \textrm{diag}(\bm{\sigma}_{i}) ).
\label{sample}
\end{eqnarray}
Since the sample operation could not provide gradient information, we leverage the reparametrization trick~\citep{kingma2NeurIPS_reference014auto} to sample the latent graph representation, i.e.,
\begin{eqnarray}
\mathbf{Z}_{\mathbb{G}} = \bm{\mu} + \epsilon \bm{\sigma}, \ \epsilon \sim \mathcal{N}(\bm{0},\bm{1}),
\label{re-para}
\end{eqnarray}
where $\epsilon$ denotes the random Gaussian noise subject to the standard normal distribution. Consequently, we can generate a negative graph set including edges and nodes by
\begin{eqnarray}
\tilde{\mathbf{A}} = \mathcal{T}(\mathbf{Z}_{\mathbb{G}}\mathbf{Z}_{\mathbb{G}}^{\top}), \ \ \tilde{\mathbf{X}} = \text{MLP}(\mathbf{Z}_{\mathbb{G}}),
\label{generation}
\end{eqnarray}
where $\text{MLP}(\cdot)$ denotes an MLP-based decoder, which aims to generate anomalous attributes $\tilde{\mathbf{X}}$ from the latent graph representations. Then the anomalous adjacent matrix $\tilde{\mathbf{A}}$ can be generated from latent graph representation learned by Edge-VGAE following $\eqref{generation}$ without the MLP-based decoder.

Our expectation is to generate high-quality anomalous graphs that closely resemble normal ones but still can be distinguished by the classifier.  This requires a high level of similarity between the generated anomaly graphs and the normal graphs, which can be regarded as minimizing the discrepancy between the generated attributes and the normal ones, with a similar objective for the generated adjacency matrix. Therefore, we propose to minimize the following discrepancy loss
\begin{eqnarray}
L_{\text{dis}} =\frac{1}{N} \sum_{i =  1}^{N} \left(\left \| \mathbf{X}_{i} -\tilde{\mathbf{X}}_{i}\right \| ^{2}_{F} - (\mathbf{A}_{i}\log(\tilde{\mathbf{A}}_{i}) + (1-\mathbf{A}_{i})\log(1-\tilde{\mathbf{A}}_{i}))\right),
\label{recon}
\end{eqnarray}
where $\tilde{\mathbf{X}}_{i}$ and $\tilde{\mathbf{A}}_{i}$ are the node attribute and adjacency matrix generated by the Node-VGAE and Edge-VGAE of $\mathcal{G}_{\phi}$ respectively, i.e., $\tilde{\mathbf{X}}_{i}, \tilde{\mathbf{A}}_{i} = \mathcal{G}_{\phi} (\mathbf{X}_{i}, \mathbf{A}_{i})$. The first term denotes the attribute reconstruction loss, and the second term denotes the binary cross-entropy loss. Additionally, the distribution of learned latent representation $\mathbf{Z}_{\mathbb{G}}$ is expected to follow a pre-defined prior distribution, which allows the generated latent representations $\mathbf{Z}_{\mathbb{G}}$ to be uniformly distributed in the latent space, ensuring the diversity of generated graphs. We can achieve this by penalizing the KL-divergence between $q(\mathbf{Z}_{\mathbb{G}}|\mathbf{H}_{\mathbb{G}})$ and a prior distribution $P(\mathbf{Z})$, i.e., $KL[q(\mathbf{Z}_{\mathbb{G}}|\mathbf{H}_{\mathbb{G}},\mathbf{A})|| P(\mathbf{Z})]$, where $P(\mathbf{Z}) = \prod_{i} p(\mathbf{Z}_{i}) = \prod_{i} \mathcal{N}(\mathbf{Z}_{i}|\mathbf{0}, \mathbf{I})$ typically follows a Gaussian prior distribution. The overall objective function of the perturbation learning-based approach is
\begin{eqnarray}
\begin{split}
\underset{\phi}{\min}~\underset{\omega}{\max} &\underset{\mathbf{X}_{i}, \mathbf{A}_{i}\sim \mathbb{P}_{\mathbb{G}}}{\mathbb{E}} [\mathcal{D}_{\omega}(\mathbf{X}_{i}, \mathbf{A}_{i})]- \underset{\mathbf{X}_{i}, \mathbf{A}_{i}\sim \mathbb{P}_{\mathbb{G}}}{\mathbb{E}}[\mathcal{D}_{\omega}(\mathcal{G}_{\phi}(\mathbf{X}_{i}, \mathbf{A}_{i}))]\\
&+ \lambda L_{\text{dis}} - \gamma KL[q(\mathbf{Z}_{\mathbb{G}}|\mathbf{H}_{\mathbb{G}},\mathbf{A})|| P(\mathbf{Z})].
\label{rec-based-function}
\end{split}
\end{eqnarray}
As the discrepancy loss and KL-divergence terms are specific to the generator, we update both of them during the training of the generator. This is a perturbation learning-based variant because the pseudo-anomalous graphs are generated via perturbing the latent variable of normal graphs. For convenience, we call this method SDGG-ATII. Compared to the GAN-based method SDGG-ATI, SDGG-ATII offers better interpretability by explicitly guiding the generator to generate pseudo-anomalous graphs that closely resemble the normal ones. Additionally, SDGG-ATII offers better control over the diversity of the generated graphs by penalizing the KL-divergence between the learned latent graph representation distribution and a prior Gaussian distribution. Compared to \eqref{eq_theta_G}, we explicitly defined the discrepancy loss $\eqref{recon}$ to guarantee the generated anomalous graphs surrounded the normal ones, and learn the decision boundary from the adversarial training of generator and discriminator. This variant offers improved interpretability in contrast to SDGG-ATI which relies solely on adversarial training between the generator and discriminator to guarantee the constraint.
Nevertheless, SDGG-ATII still suffers from the instability of the min-max optimization.

\subsection{Self-Discriminative Modeling: SDGG-NAT}\label{sec_NAT}
To address the instability of the min-max optimization in the adversarial training approaches SDGG-ATI and SDGG-ATII, we further propose a non-adversarial variant for the perturbation learning-based method, which avoids the instability problem of GANs and simplifies the training process. Specifically, rather than training a generator and a discriminator to compete against each other, we directly train a classifier $f_{\theta}$ to distinguish the anomalous graphs produced by generator $\mathcal{G}_{\phi}$ from normal ones.

We accomplish this by utilizing Node-VGAE and Edge-VGAE to produce a set of anomalous graphs $\tilde{\mathbb{G}}$ from normal graphs $\mathbb{G}$, then train a classifier to distinguish them. The overall objective is
\begin{eqnarray}
\underset{\theta, \phi}{\textrm{minimize}}\ \frac{1}{N} \sum_{i = 1}^{N}\big( \ell(y_{i},f_{\theta}(\mathbf{X}_{i}, \mathbf{A}_{i}))+\ell(\tilde{y}_{i},f_{\theta}(\mathcal{G}_{\phi}(\mathbf{X}_{i}, \mathbf{A}_{i}))\big)+ \lambda L_{\text{dis}} - \gamma KL[q(\mathbf{Z}_{\mathbb{G}}|\mathbf{H}_{\mathbb{G}},\mathbf{A})|| P(\mathbf{Z})],
\label{non-adversarial}
\end{eqnarray}
where $\ell$ denotes the binary cross-entropy loss of the classifier, $L_{\text{dis}}$ and the KL-divergence are exactly the same as \eqref{rec-based-function}. The classifier is based on GIN which receives attribute and adjacency matrices as inputs, allowing for consideration of the structural information of the graphs. Importantly, our method is unsupervised, requiring no supervised information whatsoever. We simply set $y_{1} = \cdots = y_{N} = 0$ for the normal graphs, and $\tilde{y}_{i} = \cdots = \tilde{y}_{N} =1$ for the generated anomalous graphs. Compared to \eqref{eq_theta_G}, we directly learn the decision boundary by simultaneously training a classifier with a generator that produced high-quality pseudo-anomalous graphs for the classifier. This makes our method particularly appealing for real-world applications where obtaining labeled data is challenging and costly.
The detailed training flows of the proposed SDGG-ATI, SDGG-ATII, and SDGG-NAT are given in the supplementary material due to the space limitation of the paper.

\section{Experiment}
\label{sec3}
In this section, we evaluate the proposed methods via comprehensive experiments on several molecule and social network graph datasets in comparison with state-of-the-art methods.

\subsection{Datasets, Baselines, and Experimental Settings}
\label{sec3.1}
\paragraph{Datasets.} In this paper, we experiment on different types of benchmarks, including four small molecule datasets, three biology datasets, and two social network datasets. These three types of data are typical graph-structured data in real-world scenarios. Moreover, we also consider four large molecule datasets to evaluate the anomaly detection performance in large-scale imbalanced graph datasets. Table~\ref{Dataset} briefly describes the main information of each dataset, and more details refer to the supplementary material.

\begin{wraptable}{r}{8cm}\scriptsize
\vspace{-10pt}
\centering
\caption{Brief information of the benchmarks.}
\renewcommand{\arraystretch}{0.9}
\begin{tabular}{lccccc}
\toprule
Dataset name & Graphs &  Average $[V]$&Classes &Types\\
\midrule
MUTAG        &188   &17.93    &2 &Molecule\\
AIDS         &2000  &15.69    &2 &Molecule\\
COX2         &467   &41.22    &2 &Molecule\\
ER\_MD       &446   &21.33    &2 &Molecule\\
PROTEINS     &1113  &39.06    &2 &Biology\\
DD           &1178  &284.32   &2 &Biology\\
ENZYMES      &600   &32.63    &6 &Biology\\
IMDB-Binary  &1000  &19.77    &2 &Social networks\\
COLLAB       &5000  &74.49    &3 &Social networks\\
SW-620       &40532 &26.06    &2 &Molecule\\
MOLT-4       &39765 &26.10    &2 &Molecule\\
PC-3         &27509	&26.36    &2 &Molecule\\
MCF-7        &27770 &26.40    &2 &Molecule\\
\bottomrule
\end{tabular}
\label{Dataset}
\end{wraptable}

\paragraph{Baselines.} We demonstrate the effectiveness of the proposed methods by comparison with several state-of-the-art methods including four graph kernel methods and eleven GNN-based graph-level anomaly detection approaches (to be shown in the tables of results).


\textbf{Experimental settings.}
For the proposed methods, we describe the detailed network structures, hyper-parameter settings, and training details in the supplementary material due to the space limitation. For the baseline models, we reproduced the experimental results for all of them by executing their official codes. Notably, we consider two types of experiments in this paper to evaluate anomaly detection performance. The first experiment focuses on the one-class classification task, where we respectively treat each class of a dataset as the normal class and assess the anomaly detection performance for each class individually. The second experiment involves anomaly detection on large-scale imbalanced graph datasets, where the class with a small number of samples is designated as the anomaly. We choose AUC as the evaluation metric, run each experiment 10 times and report the means and standard deviations.


\subsection{Comparison Results with State-of-the-art Approaches}\label{sec3.3}
Table~\ref{result} summarizes the performance of our methods compared to state-of-the-art approaches following the one-class classification setting (more results refer to the supplementary material). Our evaluations cover various types of graph-structured data, including molecules, biological data, and social networks. Overall, we have the following observations from the experimental results:
\begin{wrapfigure}{R}{0.4\textwidth}
    \centering
\includegraphics[width=1\linewidth,height=!]{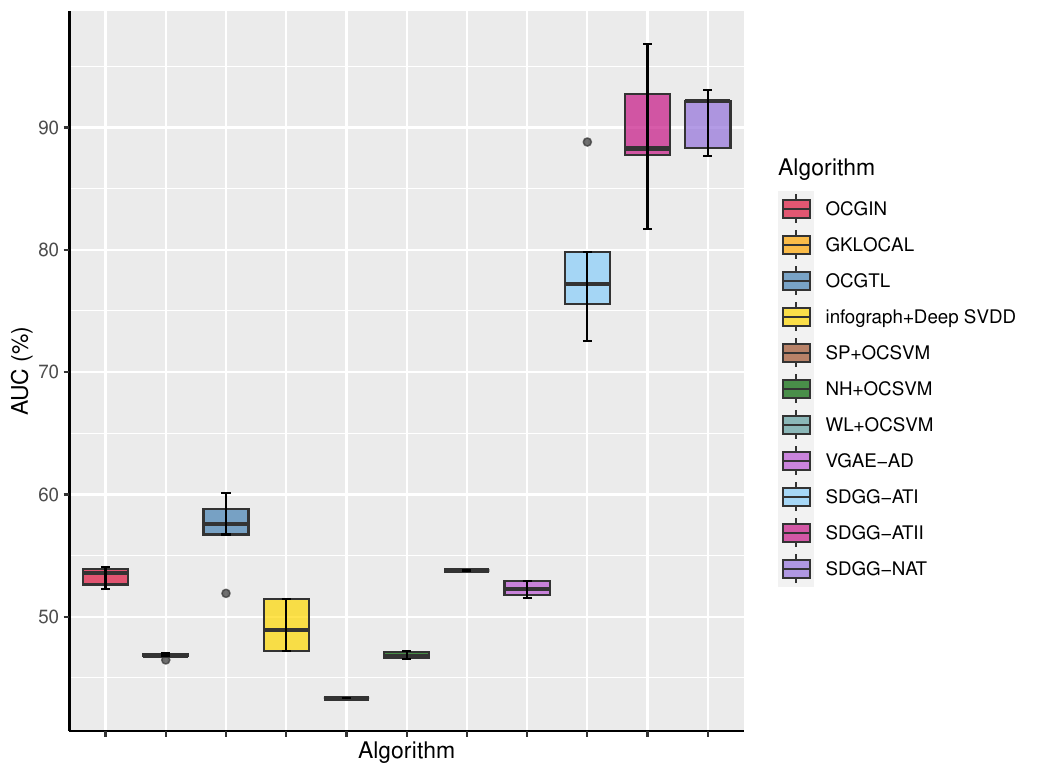}\\
    \caption{The results of multi-class AD.}
    \label{fig:multi_class_experiment}
\end{wrapfigure}
\begin{table}[h]
	\centering
	\caption{Average AUCs with standard deviation (10 trials) on MUTAG, AIDS, and PROTEINS. The best results are marked in \textbf{bold} and '--' means out of memory.}
     \resizebox{\textwidth}{!}{
	\begin{tabular}{l|c|c|c|c|c|c}
		\toprule
		\multirow{2}*{Method/Dataset}& \multicolumn{2}{c|}{MUTAG}  & \multicolumn{2}{c|}{AIDS}          & \multicolumn{2}{c}{PROTEINS}   \\ \cline{2-7}
		& \multicolumn{1}{c|}{0}             & \multicolumn{1}{c|}{1} & \multicolumn{1}{c|}{0}             & \multicolumn{1}{c|}{1}               & \multicolumn{1}{c|}{0}               & \multicolumn{1}{c}{1}                       \\ \hline
		SP~\citep{borgwardt2005shortest} & \multicolumn{1}{c|}{59.17$\pm$0.00} & 26.08$\pm$0.00 & \multicolumn{1}{c|}{97.78$\pm$0.00}                                          & \multicolumn{1}{c|}{28.32$\pm$0.00}           & \multicolumn{1}{c|}{66.83$\pm$0.00}&  52.02$\pm$0.00  \\
		WL~\citep{shervashidze2011weisfeiler}            & \multicolumn{1}{c|}{65.09$\pm$0.00} & 29.60$\pm$0.00     & \multicolumn{1}{c|}{93.81$\pm$0.00} & \multicolumn{1}{c|}{23.41$\pm$0.00} & \multicolumn{1}{c|}{73.19$\pm$0.00}   &50.19$\pm$0.00      \\
		NH~\citep{hido2009linear} & \multicolumn{1}{c|}{79.59$\pm$2.74} & 16.79$\pm$0.62  & \multicolumn{1}{c|}{96.85$\pm$0.21} & \multicolumn{1}{c|}{49.92$\pm$0.54}  &  \multicolumn{1}{c|}{68.28$\pm$0.00} &55.51$\pm$0.00 \\
		RW~\citep{vishwanathan2010graph} & \multicolumn{1}{c|}{65.03$\pm$3.12} & \multicolumn{1}{c|}{86.98$\pm$0.00} & 15.04$\pm$0.00& \multicolumn{1}{c|}{40.84$\pm$3.10}                   & \multicolumn{1}{c|}{--}                     & \multicolumn{1}{c}{--}            \\
  \midrule
  VGAE-AD~\citep{kipf2016variational}          & 70.00$\pm$4.44 & 73.30$\pm$5.40       & \multicolumn{1}{c|}{56.59$\pm$1.59 }                    & 51.11$\pm$0.34                      & \multicolumn{1}{c|}{72.22$\pm$3.70}             & 57.00$\pm$0.99 \\
		OCGIN~\citep{zhao2021using} & \multicolumn{1}{c|}{88.40$\pm$2.14} & 74.66$\pm$1.68 & \multicolumn{1}{c|}{90.65$\pm$2.04}        & \multicolumn{1}{c|}{81.52$\pm$3.76}               & \multicolumn{1}{c|}{55.01$\pm$9.65}                             & \multicolumn{1}{c}{47.77$\pm$7.64}             \\
		InfoGraph~\citep{sun2020infograph}    & \multicolumn{1}{c|}{88.05$\pm$4.48} & 61.66$\pm$20.52             & \multicolumn{1}{c|}{84.17$\pm$5.50}          & \multicolumn{1}{c|}{87.41$\pm$2.27}                             & \multicolumn{1}{c|}{65.04$\pm$13.35} &47.02$\pm$6.92                               \\
 GLocalKD~\citep{ma2022deep}         & \multicolumn{1}{c|}{50.23$\pm$23.90} & 90.59$\pm$0.61       & \multicolumn{1}{c|}{99.15$\pm$0.03}   & \multicolumn{1}{c|}{17.42$\pm$21.09} & \multicolumn{1}{c|}{72.12$\pm$0.08}  & 74.80$\pm$0.12 \\
OCGTL~\citep{qiu22raising}     & 65.70$\pm$2.10 & 75.79$\pm$22.12               & \multicolumn{1}{c|}{98.09$\pm$0.48}   & \multicolumn{1}{c|}{99.34$\pm$0.06} & \multicolumn{1}{c|}{63.20$\pm$5.40}  & 58.10$\pm$6.10 \\
\midrule
SDGG-AT\uppercase\expandafter{\romannumeral1}      & \textbf{100.00$\pm$0.00} & 98.50$\pm$2.53    & \multicolumn{1}{c|}{\textbf{100.00$\pm$0.00}} & \textbf{100.00$\pm$0.00}                  & \multicolumn{1}{c|}{90.84$\pm$0.15} & 89.19$\pm$0.17 \\
SDGG-AT\uppercase\expandafter{\romannumeral2}  & 99.31$\pm$1.42 & \textbf{99.68$\pm$2.85} & \multicolumn{1}{c|}{\textbf{100.00$\pm$0.00}} & 81.10$\pm$37.80                     & \multicolumn{1}{c|}{87.97$\pm$5.70}& 89.19$\pm$0.56 \\
SDGG-NAT       & \textbf{100.00$\pm$0.00} & 99.36$\pm$0.35     & \multicolumn{1}{c|}{99.98$\pm$0.00}  & \textbf{100.00$\pm$0.00}                     & \multicolumn{1}{c|}{\textbf{95.91$\pm$2.55}}& \textbf{96.26$\pm$0.05} \\
	\bottomrule
	\end{tabular}}
\label{result}
\vspace{-10pt}
\end{table}
\begin{itemize}[leftmargin=*]
\item Our approaches significantly outperform the graph kernels and other GNN-based methods across all datasets. For example, they outperform the closest competitor by over 20\% in terms of AUC on MUTAG and PROTEINS, and also exhibit remarkable performance on other datasets.
\item We can observe a phenomenon called ``performance flip'', where the performance of different classes in a dataset may have significant differences, in many approaches such as most graph kernels, OCGIN
, and GLocalKD on AIDS. Conversely, the ``performance flip'' is largely absent from the proposed methods, showing robust and competitive performance in each class across all benchmark datasets.
\item The variance of the performance generally reflects the stability of the model. Although graph kernels show stable performance, their overall results are not satisfactory. Furthermore, the GNN-based methods also exhibit larger variances on specific datasets. Nevertheless, in the non-adversarial version of our method, i.e., SDGG-NAT, we can observe a smaller variance in most cases, which fully demonstrates the stability of the proposed method. We also supplement the overall analysis for the proposed three variants of SDGG to further support our claim, which refers to Table~\ref{OverallAnalysis} in Appendix~\ref{appendix_results}.
\end{itemize}

Besides, we also conduct a multi-class graph-level anomaly detection experiment on ENZYMES, where multiple classes are regarded as anomalies and others as normal ones. Specifically, we set the class $\{0, 1, 2, 3\}$ as the normal classes and $\{4, 5\}$ as the anomalous classes. Figure~\ref{fig:multi_class_experiment} shows the experimental results of our methods against several state-of-the-art GNN-based GAD methods. We can observe that the proposed three methods significantly outperform all the baselines with a large margin (more than 20\%). This demonstrates the feasibility and potential of the proposed methods in dealing with multi-class GAD scenarios. Moreover, SDGG-NAT and SDGG-AT\uppercase\expandafter{\romannumeral2} achieve more outstanding performance than SDGG-AT\uppercase\expandafter{\romannumeral1}, and SDGG-NAT exhibits more stability compared with SDGG-AT\uppercase\expandafter{\romannumeral2} as it has less performance fluctuations.

\subsection{Comparison Results on Large-Scale Imbalanced Datasets}
We further evaluate the feasibility of our methods on large-scale imbalanced datasets including SW-620, MOLT-4, PC-3, and MCF-7, where we treat the rare ``active'' status in anti-cancer molecules of these datasets as anomalies. Table~\ref{latge_imbalanced} shows the experimental results of our methods compared to several state-of-the-art GNN-based approaches.
\begin{table}[h]\footnotesize
	\centering
	\caption{Average AUCs (\%) with standard deviation (\%) (10 trials) on large-scale imbalanced graph datasets. The best results are marked in \textbf{bold}.}
 \label{large}

\begin{tabular}{p{4.3cm}|cccc}
\toprule    Method/Dataset&SW-620   & MOLT-4   & PC-3  & MCF-7   \\
\midrule
\multicolumn{5}{l}{Supervised graph-level anomaly detection approaches}\\
\midrule
    GCN~\citep{kipf2017semi} & 74.90$\pm$0.74 &72.55$\pm$0.52  & 75.36$\pm$2.13  &72.70$\pm$1.05    \\
    DGCNN~\citep{zhang2018end} &80.06$\pm$0.42 & 76.50$\pm$0.60  &79.15$\pm$1.84 &76.41$\pm$0.81    \\
    GIN~\citep{xupowerful}   &78.61$\pm$2.85  &75.86$\pm$1.60    & 78.44$\pm$1.67 &69.54$\pm$1.15   \\
    SOPOOL~\citep{wang2020second}  &75.51$\pm$5.06  &75.11$\pm$0.97    &69.37$\pm$1.53  &75.64$\pm$2.17  \\
    RWGNN~\citep{nikolentzos2020random} &\multirow{2}*{73.37$\pm$0.36}  &\multirow{2}*{71.30$\pm$1.23}    &\multirow{2}*{76.27$\pm$0.86}  &\multirow{2}*{70.47$\pm$1.26}  \\
    iGAD~\citep{zhang2022dual}  &85.82$\pm$0.69  &83.59$\pm$1.07    &86.04$\pm$1.14  &83.22$\pm$0.64  \\
\midrule
\multicolumn{5}{l}{Unsupervised graph-level anomaly detection approaches}\\
\midrule
    OCGTL~\citep{qiu22raising}   &67.69$\pm$0.02 &57.42$\pm$2.38 &68.42$\pm$1.73 &64.92$\pm$1.92    \\
    GLocalKD~\citep{ma2022deep}  & 64.14$\pm$0.92 &61.43$\pm$1.26 &64.79$\pm$1.22 &  61.43$\pm$1.26    \\
\midrule
    SDGG-AT\uppercase\expandafter{\romannumeral1}    &90.19$\pm$8.94  &90.25$\pm$7.57  &91.59$\pm$6.73  &81.62$\pm$8.18    \\
    SDGG-AT\uppercase\expandafter{\romannumeral2}  &92.91$\pm$5.48  &\bf97.05$\pm$2.39  & 94.30$\pm$0.63 & 88.40$\pm$0.13\\ 
    SDGG-NAT &\bf94.26$\pm$2.86  &94.20$\pm$4.79  &\bf97.09$\pm$1.78  &\bf94.71$\pm$2.13  \\
\bottomrule
\end{tabular}
\label{latge_imbalanced}
\end{table}

Note that we compare not only with unsupervised baselines, but also with supervised ones, to demonstrate the effectiveness of our methods. From these tables, we have the following observations:
\begin{itemize}[leftmargin=*]
    \item DCGNN and iGAD outperformed OCGTL and GLocalKD, which demonstrates the usefulness of including a few anomalous graphs in the training data.
    \item SDGG-ATII and SDGG-NAT show remarkable performance across all datasets, even surpassing strong supervised baselines such as DGCNN and iGAD by more than 10\% on most datasets. The reason is that supervised methods rely heavily on real labels, which are often scarce in large-scale imbalanced datasets. Conversely, our approaches can generate high-quality pseudo-anomalous graphs. Note that supervised methods will not generalize well when the test data are not drawn from the same distribution of the training data, which occurs if the number of labeled anomalous graphs is limited. We provide a detailed explanation for this claim with a visual example in Figure~\ref{fig:explain} of Appendix~\ref{appendix_results}.
    \item Compared to SDGG-ATI, SDGG-NAT demonstrates greater stability of high accuracy, which is consistent with the motivation presented in Section \ref{sec_NAT}.
\end{itemize}

\subsection{Visualization of Learned Decision Boundary}
\begin{figure}[h!]
    \centering
    \includegraphics[height=!,width=0.8\linewidth]{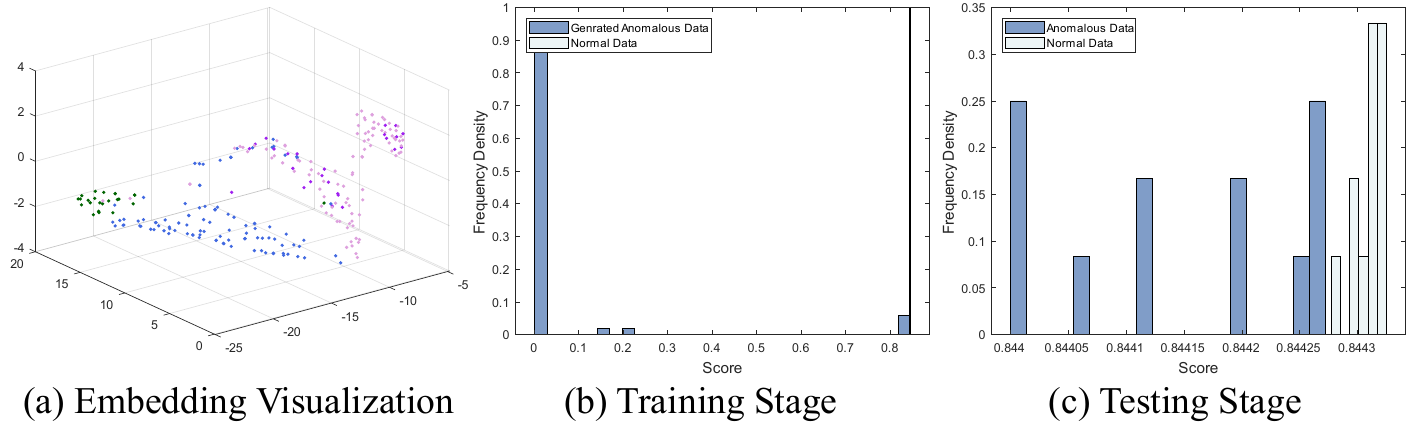}
    \caption{Visualization of SDGG-NAT on MUTAG. Note that for subfigure (a), the points marked in \textbf{\textcolor{DarkGreen}{green}}, \textbf{\textcolor{blue}{blue}}, \textbf{\textcolor{LightPink}{pink}}, and \textbf{\textcolor{DarkMagenta}{purple}} represent real anomalous data, generated anomalous data, and normal data in training and testing stages respectively. For subfigure (b), (c), the x-axis and y-axis represent the output scores and the number density of data samples within a certain interval respectively.}
    \label{fig:emb_score_visualization}
    \vspace{-8pt}
\end{figure}

We visualize the learned embeddings using t-SNE and the discriminative score of the discriminator (classifier) to intuitively demonstrate the effectiveness of the proposed methods. Figure~\ref{fig:emb_score_visualization} shows the experimental results of SDGG-NAT on MUTAG Class 1 (More results refer to the supplementary material).


From Figure~\ref{fig:emb_score_visualization} (a), we observe that the normal data from both training and testing stages approximately lie on the same manifold, while the real anomalous and generated pseudo-anomalous data are well separated into different regions from the normal data. More importantly, the generated pseudo-anomalous data interpolate between normal data and (real) anomalous data. This observation demonstrates the strong discrimination of the trained classifier, which is attributed to the high-quality pseudo-anomalous graphs generated adaptively. Moreover, the results in Figure~\ref{fig:emb_score_visualization} (b) and (c) reveal the scores of generated anomalous graphs in the classifier are significantly lower than those of the normal graphs in both the training and testing stages. This phenomenon further demonstrates that our approaches are able to accurately distinguish between normal and anomalous graphs by generating high-quality pseudo-anomalous graphs to train a powerful classifier.

\section{Conclusion}
In this paper, we proposed a novel framework for graph-level anomaly detection. The key idea is to generate pseudo-anomalous graphs that interpolate between normal graphs and (real) anomalous graphs though not presented in the training stage.  We provide three methods, namely, SDGG-ATI, SDGG-ATII, and SDGG-NAT. Particularly, SDGG-NAT has much higher learning stability and detection accuracy than the other two methods.
The comprehensive experiments on various graph benchmarks, including molecular, biological, social network, and large-scale imbalanced molecular datasets, demonstrate the effectiveness of our methods compared to state-of-the-art graph-level anomaly detection methods. Surprisingly, although our methods are unsupervised learning, they outperformed a few strong baselines of supervised learning methods for GAD. One limitation of our work is that we haven't considered any real anomalous graphs in the training stage, though they may be available in some scenarios.

\bibliography{NeurIPS_reference}
\bibliographystyle{named}

\newpage
\appendix
\onecolumn
\appendix

\section{Decision boundary visualizations of 2-D simulation}
\begin{figure}[htbp]
    \centering
    \includegraphics[height=!,width=1\linewidth]{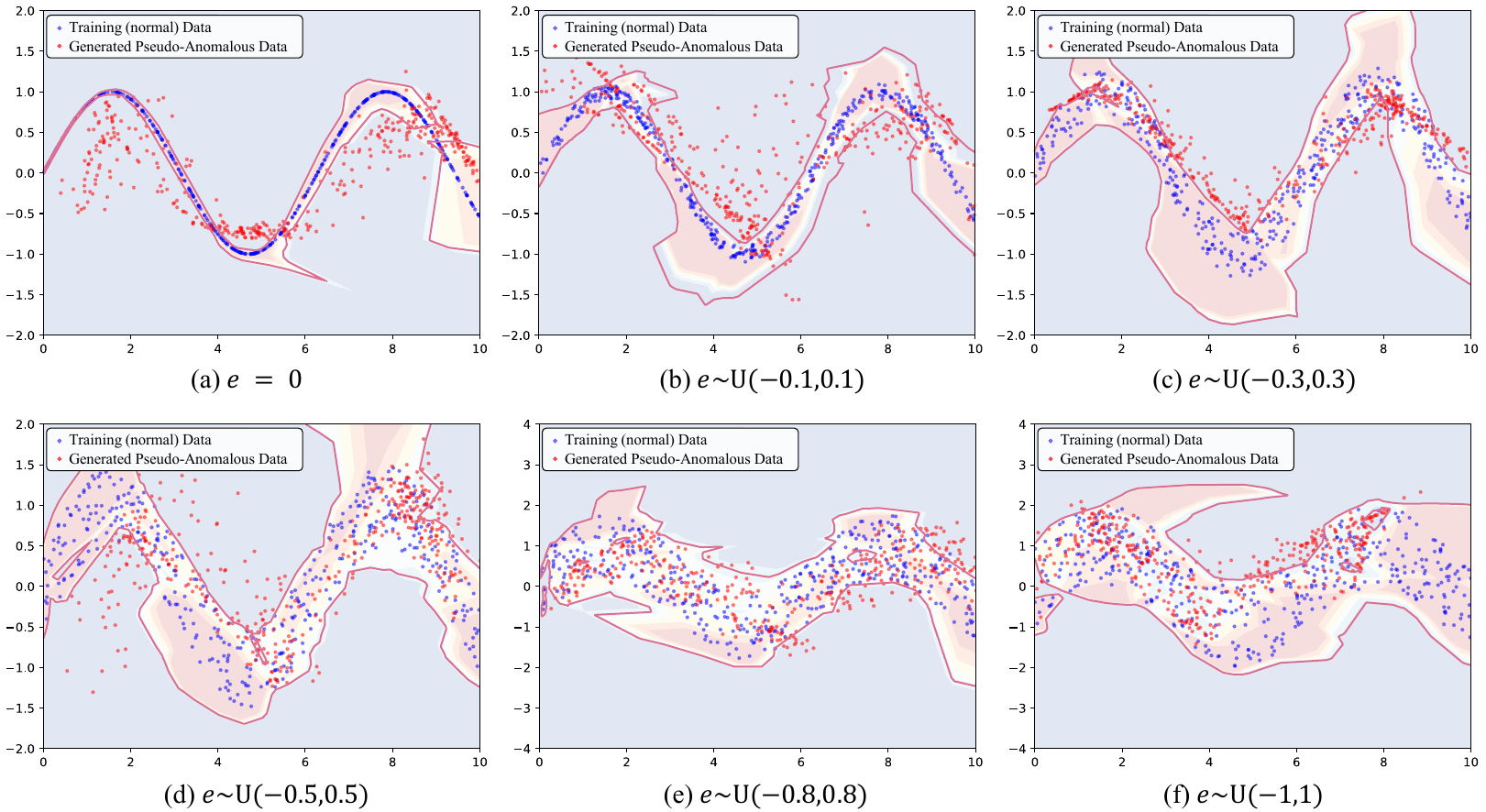}
    \vspace*{-6mm}
    \caption{The decision boundaries learned from 2-D synthetic data $x=\sin(z)+e$ with different $e$.}
    \label{fig:decision_boundary}
\end{figure}

Here we give a simulation to show the key idea and effectiveness of our methods, in addition to Figure 1 in the main paper. For convenience, we only consider SDGG-NAT and we will not use graphs because it is difficult to conduct a reasonable simulation for a number of graphs. Thus the corresponding backbones of SDGG-NAT are changed to VAE (generator) and a common MLP-based classifier.
We generate a number of synthetic 2-D samples (normal training data) using $x=\sin(z)+e$, where $e$ is a noise drawn from a uniform distribution $(-a, a)$. A larger $a$ leads to a wider normal region. In Figure \ref{fig:decision_boundary}, the pink line denotes the learned decision boundaries and the color of the figure turns from blue to red means the score given by the classifier increases. The observations are as follows.
\begin{itemize}[leftmargin=*]
    \item The generated pseudo-anomalous data usually distribute close to the training (normal) data and shows a similar manifold trend as normal one.
    \item In most cases, the classifier can distinguish those pseudo-anomalous data far from normal area.
    \item When the interval of normal data turns from narrow to wide, some generated pseudo-anomalous data may locate close to training data, but classifier would neglect most of them and draw a superior decision boundary surrounding all training data.
\end{itemize}
In conclusion, the proposed model can effectively handle normal intervals with different gaps, where the learned decision boundaries enclose the normal training data tightly. The results strongly support our assumption and motivation.

\section{Algorithm of the proposed methods}
We supplement the detailed training process of the proposed SDGG-ATI, SDGG-ATII, and SDGG-NAT in Algorithm~\ref{algorithm_1}, \ref{algorithm_2} and \ref{algorithm_3}, respectively.

\label{appendix_algorithm}
\begin{algorithm}[htbp]
	\caption{SDGG-ATI}
	\label{algorithm_1}
	\begin{algorithmic}[1]
		\REQUIRE {Input graph set $\mathbb{G}$, number of GIN layers $K$, clipping parameter $c$, learning rate $\alpha$, batch size $m$, total training epochs $\mathcal{T}$.}
		\ENSURE {The anomaly detection scores $\mathbf{s}$.}
		\STATE {Initialize the network parameters $\phi$, $\omega$;}
		\FOR{$t\rightarrow\mathcal{T}$}
            \FOR{each batch $\mathbf{G}$}
            \STATE{\textbf{Update Generator:}}
            \STATE{Unfreeze the the parameter $\phi$ of generator $\mathcal{G}_{\phi}$;}
            \STATE{Freeze the the parameter $\omega$ of discriminator $\mathcal{D}_{\omega}$;}
		\STATE{Sample random variable $\tilde{\mathbf{Z}}$ from latent Gaussian distribution $\mathbb{P}_{\tilde{\mathbf{Z}}}\sim \mathcal{N}(\bm{0}, \bm{1})$;}
  \label{step6}
		\STATE{Generate anomalous graph set $\tilde{\mathbf{G}}$ from generator $\mathcal{G}_{\phi}$ with the input $\tilde{\mathbf{Z}}$ via Eq. (3))}
  \label{step7}
        \STATE {Update the parameter $\phi$ of generator $\mathcal{G}_{\phi}$ by \\$\mathcal{G}_{\phi}\gets \nabla [-\frac{1}{m}\sum_{i=1}^{m} \mathcal{D}_{\omega}(\mathcal{G}_{\phi}(\tilde{\mathbf{Z}}_{i}), \mathcal{T}(\mathcal{G}_{\phi}(\tilde{\mathbf{Z}}_{i}) \mathcal{G}_{\phi}(\tilde{\mathbf{Z}}_{i})^{\top}))]$;\\
        $\phi \gets \phi - \alpha \cdot \textrm{RMSProp}(\phi, \mathcal{G}_{\phi}); $}
        \STATE{\textbf{Update Discriminator:}}
        \STATE{Freeze the the parameter $\phi$ of generator $\mathcal{G}_{\phi}$;}
        \STATE{Unfreeze the the parameter $\omega$ of discriminator $\mathcal{D}_{\omega}$;}
        \STATE{Repeat steps \ref{step6} - \ref{step7};}
        \STATE {Update the parameter $\omega$ of generator $\mathcal{D}_{\omega}$ by \\
        $\mathcal{D}_{\omega} \gets \nabla [- \frac{1}{m}\sum_{i=1}^{m} \mathcal{D}_{\omega}(\mathbf{X}_{i}, \mathbf{A}_{i}) + \frac{1}{m}\sum_{i=1}^{m} \mathcal{D}_{\omega}(\mathcal{G}_{\phi}(\tilde{\mathbf{Z}}_{i}), \mathcal{T}(\mathcal{G}_{\phi}(\tilde{\mathbf{Z}}_{i}) \mathcal{G}_{\phi}(\tilde{\mathbf{Z}}_{i})^{\top}))]$;\\
        $\omega \gets \omega - \alpha \cdot \textrm{RMSProp}(\omega, \mathcal{D}_{\omega});$\\
        $\omega \gets \textrm{clip}(\omega, -c,c); $
        }
		\ENDFOR
            \ENDFOR
		\STATE {Compute anomaly detection scores for test graphs via the trained discriminator $\mathcal{D}_{\omega}$;}
		\RETURN {The anomaly detection scores.}

	\end{algorithmic}
\end{algorithm}
\begin{algorithm}[htbp]
	\caption{SDGG-ATII}
	\label{algorithm_2}
	\begin{algorithmic}[1]
		\REQUIRE {Input graph set $\mathbb{G}$, number of GIN layers $K$, clipping parameter $c$, learning rate $\alpha$, batch size $m$, total training epochs $\mathcal{T}$.}
		\ENSURE {The anomaly detection scores $\mathbf{s}$.}
		\STATE {Initialize the network parameters $\phi$, $\omega$;}
		\FOR{$t\rightarrow\mathcal{T}$}
            \FOR{each batch $\mathbf{G}$}
            \STATE{\textbf{Update Generator:}}
            \STATE{Unfreeze the the parameter $\phi$ of generator $\mathcal{G}_{\phi}$;}
            \STATE{Freeze the the parameter $\omega$ of discriminator $\mathcal{D}_{\omega}$;}
		\STATE{Extrat graph-level representation with the input normal attributes $\mathbf{X}$ and adjacency matrix $\mathbf{A}$ via Eq. (4) and (5);}
  \label{step6}
		\STATE{Generate anomalous anomalous graph set $\tilde{\mathbf{G}}$ from generator $\mathcal{G}_{\phi}$ with normal attributes $\mathbf{X}$ and adjacency matrix $\mathbf{A}$ via Eq. (7), (8), (9), and (10);}
  \label{step7}
        \STATE {Update the parameter $\phi$ of generator $\mathcal{G}_{\phi}$ by \\$\mathcal{G}_{\phi}\gets \nabla [-\frac{1}{m}\sum_{i=1}^{m} \mathcal{D}_{\omega}(\mathcal{G}_{\phi}(\mathbf{X}_{i}, \mathbf{A}_{i}) + \frac{\lambda }{m} \sum_{i =1}^{m} (\Vert \mathbf{X}_{i} -\tilde{\mathbf{X}}_{i}\Vert ^{2}_{F} - (\mathbf{A}_{i}\log(\tilde{\mathbf{A}}_{i}) + (1-\mathbf{A}_{i})\log(1-\tilde{\mathbf{A}}_{i}))) -  \gamma KL[q(\mathbf{Z}_{\mathbb{G}}|\mathbf{H}_{\mathbb{G}},\mathbf{A})|| P(\mathbf{Z})]]$;\\
        $\phi \gets \phi - \alpha \cdot \textrm{RMSProp}(\phi, \mathcal{G}_{\phi}); $}
        \STATE{\textbf{Update Discriminator:}}
        \STATE{Freeze the the parameter $\phi$ of generator $\mathcal{G}_{\phi}$;}
        \STATE{Unfreeze the the parameter $\omega$ of discriminator $\mathcal{D}_{\omega}$;}
        \STATE{Repeat steps \ref{step6} - \ref{step7};}
        \STATE {Update the parameter $\omega$ of generator $\mathcal{D}_{\omega}$ by \\
        $\mathcal{D}_{\omega} \gets \nabla [- \frac{1}{m}\sum_{i=1}^{m} \mathcal{D}_{\omega}(\mathbf{X}_{i}, \mathbf{A}_{i}) + \frac{1}{m}\sum_{i=1}^{m} \mathcal{D}_{\omega}(\mathcal{G}_{\phi}(\mathbf{X}_{i}, \mathbf{A}_{i}))]$;\\
        $\omega \gets \omega - \alpha \cdot \textrm{RMSProp}(\omega, \mathcal{D}_{\omega});$\\
        $\omega \gets \textrm{clip}(\omega, -c,c); $
        }
		\ENDFOR
            \ENDFOR
		\STATE {Compute anomaly detection scores for test graphs via the trained discriminator $\mathcal{D}_{\omega}$;}
		\RETURN {The anomaly detection scores.}
	\end{algorithmic}
\end{algorithm}
\begin{algorithm}[htbp]
	\caption{SDGG-NAT}
	\label{algorithm_3}
	\begin{algorithmic}[1]
		\REQUIRE {Input graph set $\mathbb{G}$, number of GIN layers $K$, clipping parameter $c$, learning rate $\alpha$, batch size $m$, total training epochs $\mathcal{T}$.}
		\ENSURE {The anomaly detection scores $\mathbf{s}$.}
		\STATE {Initialize the network parameters $\phi$, $\theta$ for anomalous generator $\mathcal{G}_{\phi}$ and classifier $f_{\theta}$;}
		\FOR{$t\rightarrow\mathcal{T}$}
            \FOR{each batch $\mathbf{G}$}
		\STATE{Extrat graph-level representation with the input normal attributes $\mathbf{X}$ and adjacency matrix $\mathbf{A}$ via Eq. (4) and (5);}
		\STATE{Generate anomalous anomalous graph set $\tilde{\mathbf{G}}$ from generator $\mathcal{G}_{\phi}$ with normal attributes $\mathbf{X}$ and adjacency matrix $\mathbf{A}$ via Eq. (7), (8), (9), and (10);}
            \STATE {Calculate the anomalous reconstruction loss via Eq. (11)}
            \STATE {Calculate the total loss via Eq. (13)}
        \STATE {Update the parameter $\phi$ and $\theta$ of  anomalous generator $\mathcal{G}_{\phi}$ and classifier $f_{\theta}$ using backpropagation; }
		\ENDFOR
            \ENDFOR
		\STATE {Compute anomaly detection scores for test graphs via the trained classifier $f_{\theta}$;}
		\RETURN {The anomaly detection scores.}
	\end{algorithmic}
\end{algorithm}

\section{Detailed description of the datasets}
\label{appendix_dataset}
We describe more details about the datasets used in our experiment in Table~\ref{detailDataset}, which further includes the average number of edges and the node classes. Besides, Table~\ref{ImbalanceRatio} shows the imbalance ratio of large-scale graph benchmarks.

\begin{table}[htbp]
\centering
\caption{Detailed information of the graph benchmarks.}
\renewcommand{\arraystretch}{0.9}
\resizebox{\textwidth}{!}{
\begin{tabular}{lccccccc}
\toprule
Dataset name & Graphs &  Average nodes & Average edges &Node classes &Graph classes &Types\\
\midrule
Small-scale and moderate-scale datasets\\
\midrule
MUTAG        &188   &17.93  &19.79   &7  &2 &Molecule\\
AIDS         &2000  &15.69  &16.20   &38  &2 &Molecule\\
COX2         &467   &41.22  &43.45   &8  &2 &Molecule\\
ER\_MD       &446   &21.33  &234.85  &10  &2 &Molecule\\
PROTEINS     &1113  &39.06  &72.82   &3  &2 &Biology\\
DD           &1178  &284.32 &715.66  &82  &2 &Biology\\
ENZYMES      &600   &32.63  &62.14   &3  &6 &Biology\\
IMDB-Binary  &1000  &19.77  &96.53   &--  &2 &Social networks\\
COLLAB       &5000  &74.49  &2457.78 &--  &3 &Social networks\\
\midrule
Large-scale and imbalanced datasets\\
\midrule
SW-620       &40532 &26.06  &28.09   &65  &2 &Molecule\\
MOLT-4       &39765 &26.10  &28.14   &64  &2 &Molecule\\
PC-3         &27509	&26.36  &28.49   &45  &2 &Molecule\\
MCF-7        &27770 &26.40  &28.53   &46  &2 &Molecule\\
\bottomrule
\end{tabular}}
\label{detailDataset}
\end{table}

\begin{table}[htbp]
\centering
\caption{The imbalance ratio of large-scale graph benchmarks.}
\label{results4}
\renewcommand{\arraystretch}{0.6}
\begin{tabular}{l|c|c|c}
\toprule
Datasets & Class& \# Number of Graphs & Imbalance Ratio  \\
\midrule
\multirow{2}{*}{SW-620}& Normal     & 38,122  &  \multirow{2}{*}{5.95\%} \\
& Anomalous &  2,410   \\\midrule
\multirow{2}{*}{MCF-7}&   Normal    & 25,476  &  \multirow{2}{*}{8.26\%} \\
& Anomalous &  2,294   \\\midrule
\multirow{2}{*}{PC-3}&    Normal   &  25,941 &  \multirow{2}{*}{9.34\%} \\
& Anomalous & 1,568    \\\midrule
\multirow{2}{*}{MOLT-4}&  Normal     & 36,625  &  \multirow{2}{*}{7.90\%} \\
& Anomalous &  3,140    \\
\bottomrule
\end{tabular}
\label{ImbalanceRatio}
\end{table}

\section{Detailed experimental settings}
\label{appendix_experimental_setting}
We supplement more details of the experimental settings in the paper, which includes the network architecture of the proposed methods, settings of trade-off parameters, training details, data split, baseline settings, etc.
\begin{itemize}
    \item \textbf{Network architecture:}  For the network architecture of the proposed SDGG-ATI, SDGG-ATII, and SDGG-NAT, we utilize a 3-layer GIN as the backbone network for the generator and discriminator (classifier), except the generator of SDGG-ATI which is an MLP-based neural network. The aggregated dimension and the latent dimension in our method are set to 16 and 10, respectively. The source code is also included in the supplementary material to ensure the reproducibility of our methods.
    \item \textbf{Trade-off parameters:} The coefficient of anomalous reconstruction loss $\lambda$ varies in $\{0.01,0.1,1,10\}$. The specific value is chosen according to resist the influence of classifier loss. The other parameter $\beta$ of KL-divergence loss is 1e-5 and the clip value of the adversarial loss is fixed by 0.01. Furthermore, we further assess the impact of variations in the values of $\lambda$ and $\beta$ on performance in Appendix~\ref{appendix_parameter}.
    \item \textbf{Training details:} For small-scale graph datasets, we utilize a fixed batch size of 4, while we increase the batch size to 256 to accommodate the requirement of experiment on larger-scale datasets. Besides, we set the learning rate $\alpha$ to 0.001 and the total training epochs to 300, utilizing RMSprop~\citep{tieleman2012lecture} optimizer for SDGG-ATI and SDGG-ATII, and Adam~\citep{kingma2014adam} for SDGG-NAT.
    \item \textbf{Data split:} For small and moderate scale datasets, we allocate 80\% of the data from the normal class for training, and subsequently construct the testing data by combining the retained normal data with an equal or smaller number of anomalous data samples. For large-scale imbalanced datasets, we allocate 80\% of the data in the normal class as the training set, and form the test set with the rest of the normal data and all the abnormal data.
    \item \textbf{Baseline settings:} Particularly, we utilize the one-class support vector machine (OCSVM) to achieve anomaly detection for all graph kernel baselines and InfoGraph. For other baselines, we follow the settings in their papers and report the reproduced results. Note that we select the best results achieved throughout the training epochs for all algorithms to ensure a fair comparison.
    \item \textbf{Implementation:} Note we leverage PyTorch Geometric~\citep{fey2019fast} for implementation, and all experiments are executed on NVIDIA Tesla A100 GPU with AMD EPYC 7532 CPU.
\end{itemize}

\section{More experimental results}
\label{appendix_results}
In this section, we present additional experimental results for the one-class classification tasks. Tables~\ref{result-1} and \ref{result-2} show the anomaly detection results for the remaining experimental datasets. Notably, the proposed methods, SDGG-ATI, SDGG-ATII, and SDGG-NAT, consistently outperform other baselines by a significant margin, underscoring the superiority of our approaches. Besides, we evaluate the anomaly detection performance on the multi-class graph dataset ENZYMES, which contains 6 classes in total. Table~\ref{multi-class-result} presents the experimental results for each class. Our methods consistently demonstrate remarkable effectiveness across all classes of ENZYMES, surpassing all baselines by more than 20\%. Table~\ref{OverallAnalysis} summarizes the overall analysis of SDGG-ATI, SDGG-ATII, and SDGG-NAT across all benchmarks. It is evident that, in the majority of cases, SDGG-NAT achieves the best AUCs. Additionally, SDGG-NAT consistently maintains $\text{std}\leq 5\%$ across 24 out of 27 cases, and it exhibits the smallest standard deviation compared to SDGG-ATI/ATII in more cases. This provides strong evidence for the stability of SDGG-NAT.

To further demonstrate the effectiveness of the proposed SDGG methods, we also evaluate the proposed method with different metrics, e.g., F1 score, recall, and AUCPR. Table~\ref{MoreMetrics} shows the experimental results of several comparative methods on four graph benchmarks. We can observe that the three variants of the proposed SDGG exhibit superiority compared to other baseline methods across all metrics, and SDGG-NAT also outperforms SDGG-ATI and SDGG-ATII in most cases.

\begin{figure}[htbp]
  \centering
  \includegraphics[width=0.7\linewidth]{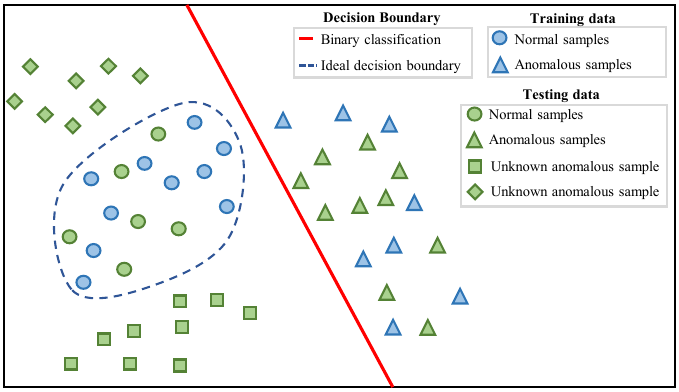}
  \caption{The illustration of the idea of SDGG. The \textbf{\textcolor{red}{red}} line denotes the decision boundary learned by the binary classification, whereas the \textbf{\textcolor{blue}{blue}} dashed line denotes the ideal decision boundary.}
    \label{fig:explain}
\end{figure}

Figure~\ref{fig:explain} offers a visual illustration of why the SDGG can outperform other supervised methods in the context of large-scale imbalanced benchmarks. This figure demonstrates a binary classification scenario where the anomalous data used for training lies on the right side of the normal data, and the trained classifier successfully categorizes them with a red decision boundary. However, there may be unknown anomalies located on the left side of the decision boundary (shown by the blue dashed line), where the binary classifier fails to detect them. This phenomenon is common in large-scale unbalanced anomaly detection due to limited supervised information, where the distribution of test data is not exactly the same as that of the training data.

\begin{table*}[htbp]
	\centering
	\caption{Average AUCs with standard deviation (10 trials) of different graph-level anomaly detection algorithms. We assess models by regarding every data class as normal data, respectively. The best results are highlighted in \textbf{bold} and ``--'' means out of memory.}
     \resizebox{\textwidth}{!}{
	\begin{tabular}{l|c|c|c|c|c|c}
		\toprule
		\multirow{2}*{Method/Dataset} & \multicolumn{2}{c|}{COX2}       & \multicolumn{2}{c|}{ER\_MD}
    & \multicolumn{2}{c}{DD}
  \\
  \cline{2-7}
& \multicolumn{1}{c|}{0}& \multicolumn{1}{c|}{1}
& \multicolumn{1}{c|}{0}& \multicolumn{1}{c|}{1}
& \multicolumn{1}{c|}{0}& \multicolumn{1}{c}{1}
\\ \hline
SP~\citep{borgwardt2005shortest}  & \multicolumn{1}{c|}{54.08$\pm$0.00} & 57.60$\pm$0.00 & \multicolumn{1}{c|}{40.92$\pm$0.00} & 38.24$\pm$0.00 & \multicolumn{1}{c|}{68.56$\pm$0.00} & 44.74$\pm$0.00 \\
WL~\citep{shervashidze2011weisfeiler}  & \multicolumn{1}{c|}{59.90$\pm$0.00} & 50.57$\pm$0.00 & \multicolumn{1}{c|}{45.71$\pm$0.00} & 32.62$\pm$0.00  & \multicolumn{1}{c|}{73.97$\pm$0.00} & 49.46$\pm$0.00  \\
NH~\citep{hido2009linear}  & \multicolumn{1}{c|}{48.41$\pm$0.00} & 47.17$\pm$0.00 & \multicolumn{1}{c|}{51.55$\pm$2.00} & 36.48$\pm$0.00 & \multicolumn{1}{c|}{74.24$\pm$0.00} & 36.84$\pm$0.00  \\
RW~\citep{vishwanathan2010graph}& \multicolumn{1}{c|}{52.43$\pm$0.00} & 65.53$\pm$0.00 & \multicolumn{1}{c|}{48.20$\pm$0.00} & 34.84$\pm$0.00  & \multicolumn{1}{c|}{--} & \multicolumn{1}{c}{--} \\
\midrule
OCGIN~\citep{zhao2021using} & \multicolumn{1}{c|}{59.64$\pm$5.78} & 56.83$\pm$7.68 & \multicolumn{1}{c|}{72.20$\pm$0.16} & 70.08$\pm$0.56 & \multicolumn{1}{c|}{66.59$\pm$4.44} & 60.03$\pm$5.34 \\
InfoGraph~\citep{sun2020infograph}  & \multicolumn{1}{c|}{48.25$\pm$6.24} & 50.29$\pm$7.00 & \multicolumn{1}{c|}{53.12$\pm$15.45} & 56.82$\pm$7.04 & \multicolumn{1}{c|}{39.42$\pm$4.36} & 64.84$\pm$2.36\\
GLocalKD~\citep{ma2022deep} & \multicolumn{1}{c|}{51.42$\pm$0.66} & 65.79$\pm$0.98 & \multicolumn{1}{c|}{57.81$\pm$17.90} &71.54$\pm$0.00 & \multicolumn{1}{c|}{19.52$\pm$0.00} & 22.03$\pm$0.01\\
OCGTL~\citep{qiu22raising}   & \multicolumn{1}{c|}{55.41$\pm$3.20} & 48.62$\pm$2.24 & \multicolumn{1}{c|}{27.55$\pm$3.17} & 69.15$\pm$2.07 & \multicolumn{1}{c|}{69.90$\pm$2.60} & 67.67$\pm$2.80\\
VGAE-AD~\citep{kipf2016variational}  & \multicolumn{1}{c|}{59.28$\pm$1.55} & 73.33$\pm$1.48 & \multicolumn{1}{c|}{59.89$\pm$7.08}& 59.48$\pm$6.94 & 51.57$\pm$0.94 & 64.95$\pm$5.34  \\ \midrule
SDGG-AT\uppercase\expandafter{\romannumeral1}   & \multicolumn{1}{c|}{71.00$\pm$8.67} & \multicolumn{1}{c|}{79.45$\pm$5.82} & \multicolumn{1}{c|}{90.52$\pm$2.93} & \multicolumn{1}{c}{95.05$\pm$1.48} & 80.87$\pm$1.76 & 88.50$\pm$3.22\\
SDGG-AT\uppercase\expandafter{\romannumeral2}  &\multicolumn{1}{c|}{68.56$\pm$5.59} & 91.78$\pm$3.95 & \multicolumn{1}{c|}{88.10$\pm$0.53}&93.61$\pm$1.76 & \textbf{91.73$\pm$0.74} & 79.09$\pm$26.38    \\
SDGG-NAT            & \multicolumn{1}{c|}{\textbf{73.91$\pm$11.52}}&\bf97.05$\pm$1.62
& \multicolumn{1}{c|}{\textbf{98.74$\pm$1.59}}         & \textbf{96.67$\pm$1.67}     & 90.71$\pm$1.17 & \textbf{97.71$\pm$1.76}
  \\

	\bottomrule
	\end{tabular}}
\label{result-1}
\end{table*}

\begin{table*}[htbp]
	\centering
 \caption{Average AUCs with standard deviation (10 trials) of different graph-level anomaly detection algorithms. We assess models by regarding every data class as normal data, respectively. The best results are highlighted in \textbf{bold} and ``--'' means out of memory.}
	\resizebox{\textwidth}{!}{
	\begin{tabular}{l|c|c|c|c|c}
		\toprule
	         \multirow{2}*{Method/Dataset} & \multicolumn{2}{c|}{IMDB-Binary} & \multicolumn{3}{c}{COLLAB}    \\
  \cline{2-6}
 & \multicolumn{1}{c|}{0}  & \multicolumn{1}{c|}{1}        & \multicolumn{1}{c|}{0}  & \multicolumn{1}{c}{1} & \multicolumn{1}{c}{2} \\ \hline
SP~\citep{borgwardt2005shortest} & \multicolumn{1}{c|}{45.92$\pm$0.00} & 47.16$\pm$0.00 & \multicolumn{1}{c|}{59.10$\pm$0.00} & \multicolumn{1}{c|}{83.97$\pm$0.00} & 79.02$\pm$0.00 \\
WL~\citep{shervashidze2011weisfeiler}  & \multicolumn{1}{c|}{51.57$\pm$0.00} & 46.07$\pm$0.00 & \multicolumn{1}{c|}{51.22$\pm$0.00} & \multicolumn{1}{c|}{80.54$\pm$0.00} & 79.96$\pm$0.00\\
NH~\citep{hido2009linear} & \multicolumn{1}{c|}{53.21$\pm$0.00} & 46.52$\pm$0.00  & \multicolumn{1}{c|}{59.76$\pm$0.00} & \multicolumn{1}{c|}{80.54$\pm$0.00} & 64.14$\pm$0.00\\
RW~\citep{vishwanathan2010graph} & \multicolumn{1}{c|}{49.51$\pm$0.00} & 53.11$\pm$0.00 & \multicolumn{1}{c|}{--} & \multicolumn{1}{c|}{--} & \multicolumn{1}{c}{--} \\
\midrule
OCGIN~\citep{zhao2021using}
 & 40.47$\pm$10.83 & 44.22$\pm$4.99 & \multicolumn{1}{c|}{42.17$\pm$6.06} & \multicolumn{1}{c|}{75.65$\pm$20.35} & 19.06$\pm$8.57\\
InfoGraph~\citep{sun2020infograph}
& 63.53$\pm$2.77 & 58.36$\pm$9.95 & \multicolumn{1}{c|}{56.62$\pm$5.97} & \multicolumn{1}{c|}{79.26$\pm$9.86} & 40.62$\pm$9.78 \\
GLocalKD~\citep{ma2022deep}
 & 53.83$\pm$1.24 & 53.34$\pm$0.06 & \multicolumn{1}{c|}{46.38$\pm$0.03} & \multicolumn{1}{c|}{50.16$\pm$0.20} & 52.98$\pm$0.04 \\
OCGTL~\citep{qiu22raising}
& 65.10$\pm$1.80 & 64.12$\pm$1.27 & \multicolumn{1}{c|}{65.04$\pm$4.33} & 89.08$\pm$2.39 & 40.29$\pm$5.41 \\
VGAE-AD~\citep{kipf2016variational}  & 65.36$\pm$0.78 & 67.22$\pm$3.49 & 50.96$\pm$0.04 & --&-- \\
\midrule
SDGG-AT\uppercase\expandafter{\romannumeral1}  & 62.92$\pm$0.62 & 86.53$\pm$0.00 & \multicolumn{1}{c|}{65.85$\pm$7.20} & \multicolumn{1}{c|}{82.82$\pm$0.32} & 73.57$\pm$7.55\\
SDGG-AT\uppercase\expandafter{\romannumeral2}  & 90.51$\pm$2.79 & 87.93$\pm$0.25 & \multicolumn{1}{c|}{54.49$\pm$6.79}& \multicolumn{1}{c|}{82.95$\pm$0.45}&78.68$\pm$0.43 \\
SDGG-NAT  & \bf93.37$\pm$1.57 & \textbf{93.07$\pm$1.54}& \multicolumn{1}{c|}{\bf87.99$\pm$6.21}       & \multicolumn{1}{c|}{\bf92.82$\pm$3.18}              & \bf94.74$\pm$1.40  \\
		\bottomrule
	\end{tabular}}
\label{result-2}
\end{table*}

\begin{table*}[htbp]
\centering
\caption{Average AUCs with standard deviation (5 trials) of the proposed models on ENZYMES dataset. The best results are highlighted in \textbf{bold} and ``--'' means out of memory.}
\resizebox{\textwidth}{!}{
\begin{tabular}{l|c|c|c|c|c|c}
\toprule
\multicolumn{1}{l|}{\multirow{2}*{Method/Dataset}} & \multicolumn{6}{c}{ENZYMES}                                                                                \\\cmidrule{2-7}
\multicolumn{1}{c|}{}    & 0               & 1               & 2               & 3               & 4                & 5               \\\midrule
SP~\citep{borgwardt2005shortest} & 58.30$\pm$0.00 & 49.50$\pm$0.00 & 48.75$\pm$0.00 & 57.55$\pm$0.00 & 65.20$\pm$0.00 & 63.45$\pm$0.00 \\
WL~\citep{shervashidze2011weisfeiler} & 56.50$\pm$0.00 & 49.70$\pm$0.00 & 60.05$\pm$0.00 & 54.10$\pm$0.00 & 44.85$\pm$0.00 & 50.85$\pm$0.00 \\
NH~\citep{hido2009linear} & 58.09$\pm$1.21 & 52.70$\pm$0.33 & 55.93$\pm$1.03 & 56.70$\pm$0.65 & 44.28$\pm$0.85 & 66.79$\pm$0.53 \\
RW~\citep{vishwanathan2010graph} & -- & -- & -- & -- & -- & -- \\
\midrule
OCGIN~\citep{zhao2021using} & 56.68$\pm$3.53 & 67.43$\pm$4.25 & 62.18$\pm$2.93 & 44.74$\pm$5.11 & 53.88$\pm$1.17 & 62.95$\pm$7.16 \\
InfoGraph~\citep{sun2020infograph} & 71.70$\pm$0.00 & 52.67$\pm$4.78 & 50.90$\pm$2.01 & 71.10$\pm$0.00 & 46.02$\pm$2.91 & 55.23$\pm$1.34 \\
GLocalKD~\citep{ma2022deep} & 58.27$\pm$0.57 & 63.44$\pm$0.11 & 53.13$\pm$0.06 & 56.50$\pm$0.04 & 59.23$\pm$0.11 & 63.08$\pm$0.06 \\
OCGTL~\citep{qiu22raising} & 61.58$\pm$2.13 & 55.04$\pm$1.34 & 46.10$\pm$0.32 & 61.74$\pm$1.59 & 62.28$\pm$2.29 & 56.92$\pm$2.72 \\
VGAE-AD~\citep{kipf2016variational} & 58.38$\pm$0.67 & 57.79$\pm$1.38 & 56.39$\pm$3.26 & -- & -- & -- \\
\midrule
SDGG-AT\uppercase\expandafter{\romannumeral1}  & 86.93$\pm$5.36 & 77.71$\pm$1.78 & 75.52$\pm$4.12 & 92.04$\pm$2.09 & 75.91$\pm$1.77 & 85.16$\pm$3.54 \\
SDGG-AT\uppercase\expandafter{\romannumeral2}  & 81.42$\pm$3.02 & 72.53$\pm$3.71 & 67.39$\pm$6.51 & 90.73$\pm$2.95 & 79.85$\pm$7.84 & 76.67$\pm$2.49 \\
SDGG-NAT  & \bf89.28$\pm$3.13 & \bf90.42$\pm$3.55 & \bf79.70$\pm$2.65 & \bf95.29$\pm$1.35 & \bf87.51$\pm$1.24 & \bf89.60$\pm$8.02 \\
\bottomrule
\end{tabular}}
\label{multi-class-result}
\end{table*}

\begin{table}[htbp]
\centering
	\caption{Average F1-score, Recall and AUCPR (\%) with standard deviation (\%) (10 trials) on large-scale imbalanced graph datasets. The best results are marked in \textbf{bold}.}
\setlength{\tabcolsep}{2mm}
 \renewcommand{\arraystretch}{0.6}
\begin{tabular}{l|c|c|c|c|c}
\toprule
Metric    &Method/Dataset   & PC-3  & MCF-7 & PROTEINS (0) &  PROTEINS (1) \\
\midrule
\multirow{5}{*}{F1-score}    &OCGTL~\citep{qiu22raising}    & 62.33$\pm$0.46& 79.34$\pm$0.31  &66.50$\pm$0.00& 66.42$\pm$0.00\\
    &GLocalKD~\citep{ma2022deep}   & 40.78$\pm$0.14& 68.03$\pm$0.02 & 66.73$\pm$5.41& 68.25$\pm$0.00\\
    \cmidrule{2-6}
    &SDGG-AT\uppercase\expandafter{\romannumeral1}    & 88.38$\pm$0.13& 82.61$\pm$0.00  &64.68$\pm$5.19 & 71.43$\pm$0.00 \\
    &SDGG-AT\uppercase\expandafter{\romannumeral2}  & 88.56$\pm$0.13&\bf83.06$\pm$0.12& 71.46$\pm$2.43& 70.83$\pm$0.12\\
    &SDGG-NAT & \bf93.36$\pm$2.30& 82.92$\pm$0.10 & \bf71.93$\pm$0.16& \bf71.52$\pm$0.13\\\midrule
    \multirow{5}{*}{Recall}    &OCGTL~\citep{qiu22raising}    & 87.22$\pm$0.64& 93.63$\pm$0.36 & \bf100.00$\pm$0.00& \bf100.00$\pm$0.00  \\
    &GLocalKD~\citep{ma2022deep}  &93.24$\pm$0.06 & 85.85$\pm$0.02  & 93.18$\pm$7.50&  95.56$\pm$0.00\\
    \cmidrule{2-6}
    &SDGG-AT\uppercase\expandafter{\romannumeral1}    &99.06$\pm$0.14 & 97.48$\pm$0.05  &90.40$\pm$7.25& \bf100.00$\pm$0.00 \\
    &SDGG-AT\uppercase\expandafter{\romannumeral2}  & \bf99.26$\pm$0.03&98.02$\pm$0.14& 97.47$\pm$2.50& \bf100.00$\pm$0.00\\
    &SDGG-NAT & 99.23$\pm$0.11& \bf99.41$\pm$0.06 & \bf100.00$\pm$0.00&\bf100.00$\pm$0.00 \\\midrule
    \multirow{5}{*}{AUCPR}    &OCGTL~\citep{qiu22raising}    & 47.46$\pm$0.78& 64.98$\pm$2.54 & 75.00$\pm$0.00& 75.00$\pm$0.00 \\
    &GLocalKD~\citep{ma2022deep}  &41.33$\pm$0.02 & 46.62$\pm$0.00  &  62.30$\pm$19.24& 82.12$\pm$0.13\\
    \cmidrule{2-6}
    &SDGG-AT\uppercase\expandafter{\romannumeral1}    & 95.19$\pm$0.63& 89.69$\pm$1.14 & 75.67$\pm$27.48 & 96.45$\pm$0.04\\
    &SDGG-AT\uppercase\expandafter{\romannumeral2}  & 95.69$\pm$0.27& 90.03$\pm$0.05& 92.78$\pm$7.31& 96.14$\pm$0.01\\
    &SDGG-NAT & \bf95.87$\pm$0.12& \bf90.87$\pm$0.25& \bf97.67$\pm$0.36& \bf96.58$\pm$0.07 \\
\bottomrule
\end{tabular}
\label{MoreMetrics}
\end{table}

\begin{table}[htbp]
\caption{Overall analysis for the three variants of SDGG across all graph benchmarks.}
\centering
\begin{tabular}{c|ccc}
\toprule
   & SDGG-ATI & SDGG-ATII & SDGG-NAT\\
\midrule
Min std & 9/27 & 10/27 &12/27\\
std$\leq$5\% & 18/27 & 19/27 & 24/27\\
Best AUC & 3/27 & 4/27 & 23/27\\
\bottomrule
\end{tabular}
\label{OverallAnalysis}
\end{table}

\section{More visualization results}
\label{appendix_visualization}
\begin{figure}[htbp]
    \centering
    \includegraphics[height=!,width=1\linewidth]{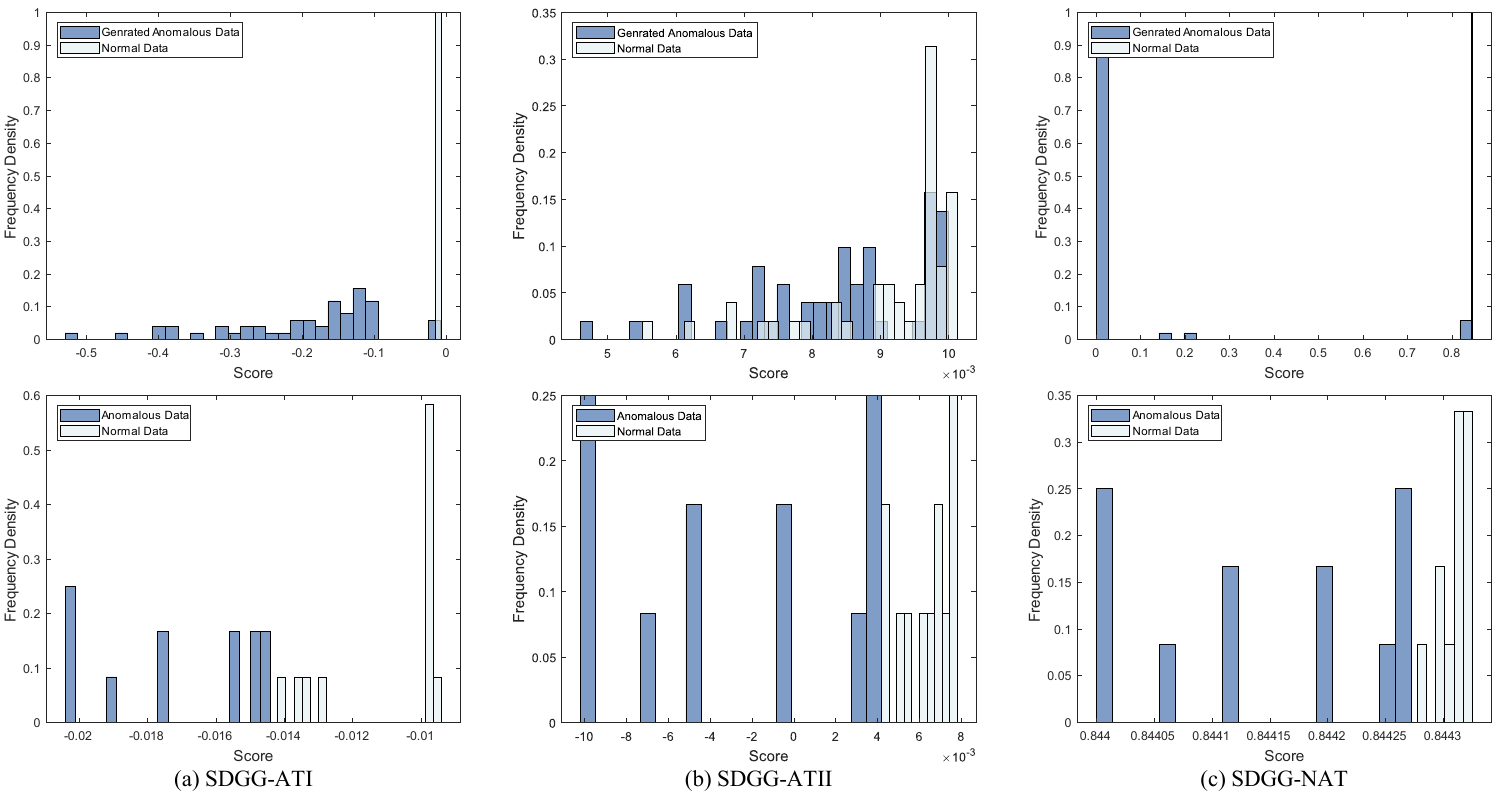}
    \vspace*{-6mm}
    \caption{The discriminative score visualizations of the proposed models on MUTAG Class 1. The top row is the result of training stage and the bottom row is that of testing stage.}
    \label{fig:output_visualization}
\end{figure}

In this section, we provide additional visualization results of our methods. Figure~\ref{fig:output_visualization} illustrates the discriminative scores of SDGG-ATI, SDGG-ATII, and SDGG-NAT on MUTAG. The top row represents the discriminative scores in the training stage, while the bottom row corresponds to the testing stage. It is evident that during the training phase, our methods effectively differentiate between the generated anomalous data and normal data, and this distinction carries over to the testing phase. These findings validate that the classifier trained using high-quality generated anomalous graphs can identify outstanding decision boundaries and exhibits excellent generalization capabilities during the testing stage. Particularly, despite observing score overlap between the generated anomaly and normal graphs during the training phase of SDGG-ATII, significant differentiation is still achieved during the testing phase. This can be attributed to our objective of training a powerful classifier by generating high-quality anomaly graphs that closely resemble normal graphs. Although the classifier may not separate these anomalies adequately during training. This may possibly be due to the over-idealization of the generated anomaly data, the learned decision boundaries are sufficiently effective in distinguishing the anomalies during the test phase.

\begin{figure}[htbp]
    \centering
    \includegraphics[height=!,width=1\linewidth]{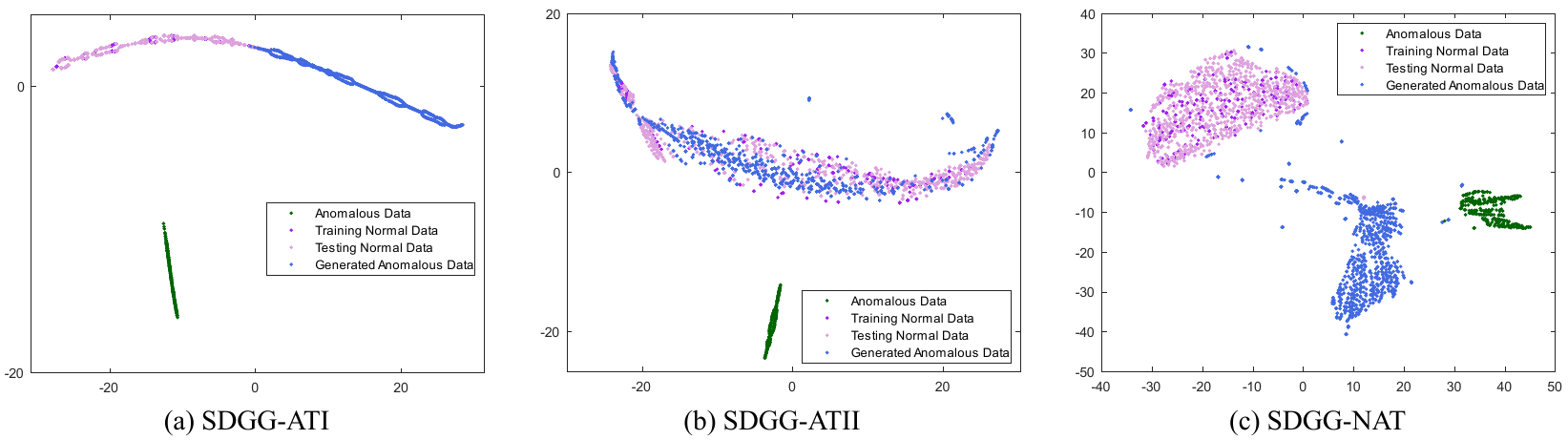}
    \caption{The 2-D t-SNE visualizations of the proposed models on AIDS Class 1.}
    \label{fig:2d_tsne}
\end{figure}

\begin{figure}[htbp]
    \centering
    \includegraphics[height=!,width=1\linewidth]{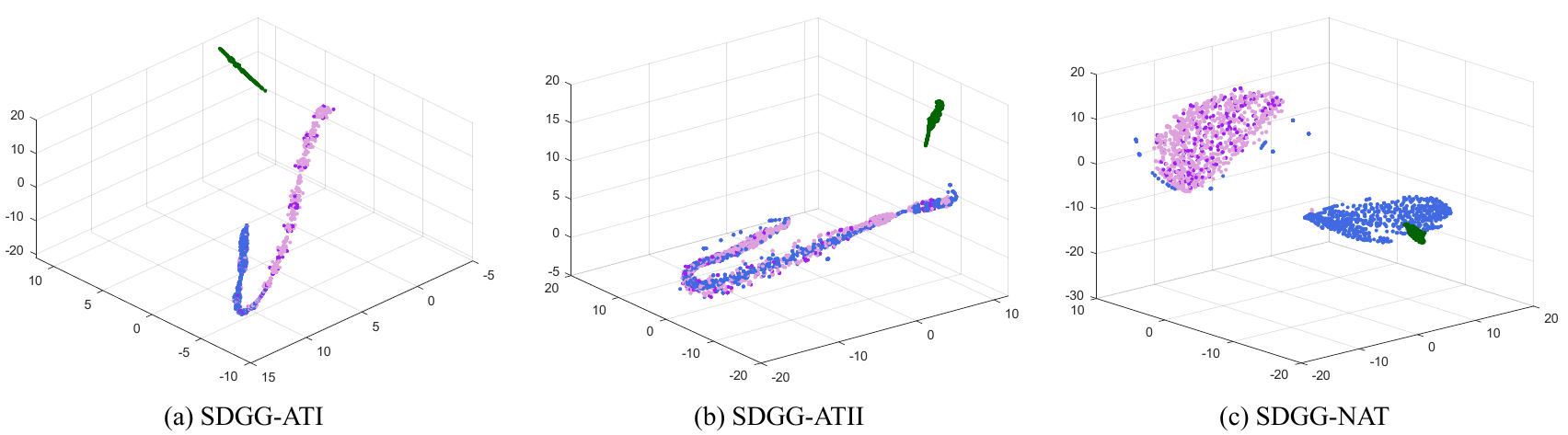}
    \caption{The 3-D t-SNE visualizations of the proposed models on AIDS Class 1. The legend is set the same as Figure \ref{fig:2d_tsne}.}
    \label{fig:3d_tsne}
\end{figure}

Additionally, we present the 2-D and 3-D t-SNE visualizations (Figure~\ref{fig:2d_tsne} and \ref{fig:3d_tsne}) to provide a comprehensive assessment of the effectiveness of the proposed SDGG-ATI, SDGG-ATII, and SDGG-NAT. These visualizations offer compelling insights into the learned decision boundaries derived from the generated anomalous data. Examining these visualizations in detail, we can observe that anomalous data, distinctly highlighted in green, are conspicuously separated from the remaining data points. This distinct separation serves as compelling evidence of the discriminative power embedded within our methods. Through these insightful visualizations, we gain a deeper understanding of the proposed methods, vividly illustrating their ability to learn effective decision boundaries and unveil intricate patterns and anomalies hidden within complex graph structures.

\section{Parameter sensitivity analysis}
\label{appendix_parameter}
\begin{figure}[htbp]
    \centering
    \includegraphics[height=!,width=0.85\linewidth]{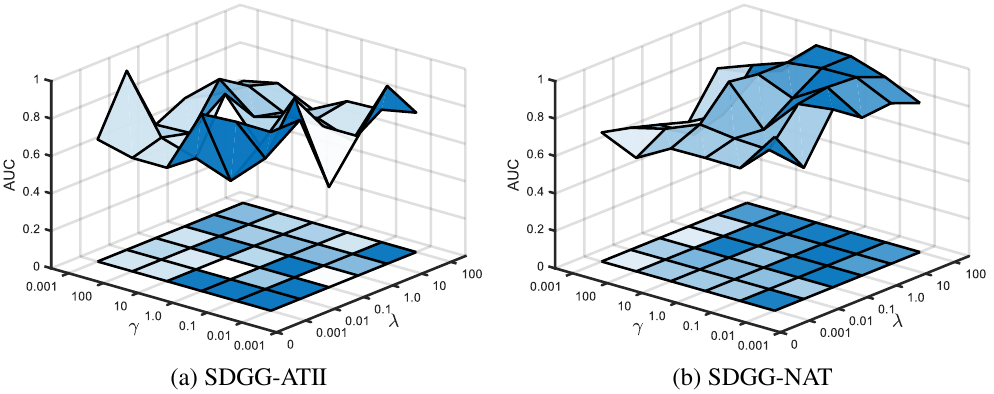}
    \caption{Parameter sensitivity analysis of SDGG-AT\uppercase\expandafter{\romannumeral2} and SDGG-NAT on COX2. $\lambda$ and $\gamma$ changes in the range of $[0.001, 100]$.}
    \label{fig:parameteranalysis}
\end{figure}

We investigate the impact of two main hyper-parameters, $\lambda$ and $\gamma$, in SDGG-AT\uppercase\expandafter{\romannumeral2} and SDGG-NAT on the anomaly detection performance. Note that SDGG-AT\uppercase\expandafter{\romannumeral1} is not included in this analysis because its loss function does not have any hyper-parameter. Specifically, we set the range of value for $\lambda$ and $\gamma$ from $0.001$ to $100$ and evaluate their influence on COX2. Figure~\ref{fig:parameteranalysis} shows the experimental results, where we have the following observations.
First, we find that a balanced trade-off of $\lambda$ and $\gamma$ is crucial for achieving ideal performance in SDGG-AT\uppercase\expandafter{\romannumeral2} and SDGG-NAT.
Either too large or too small values will generally lead to sub-optimal results.
Second, both SDGG-AT\uppercase\expandafter{\romannumeral2} and SDGG-NAT exhibit relatively stable performance across a wide range of $\lambda$ and $\gamma$ values, which demonstrates the effectiveness of our methods.
Third, we can observe that the two hyper-parameters cause less significant influence on the performance of SDGG-NAT than SDGG-AT\uppercase\expandafter{\romannumeral2}.
This further demonstrates that the non-adversarial variant of the proposed method exhibits greater stability and robustness.


\section{Comparison between VGAE-based and GIN-based backbones}
\label{appendix_backbone}
\begin{figure}[htbp]
\centering
\includegraphics[width=0.6\linewidth]{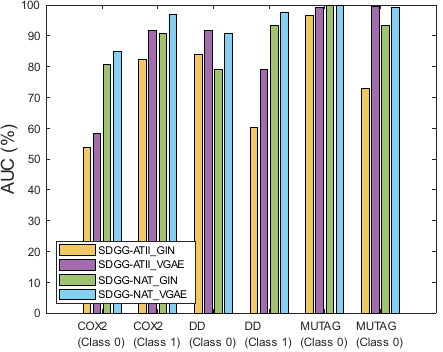}
	\caption{Comparison between VGAE-based and GIN-based generator.} \label{fig:backbone_comparison}
\end{figure}

In this section, we conduct a thorough comparison between VGAE-based and GIN-based generators to elucidate our rationale for choosing VGAE as the preferred backbone for generators in our methods. It should be noted that the key difference between VGAE-based and GIN-based backbones lies in the incorporation of variational inference that introduces stochasticity in generating anomalous graphs.
Figure \ref{fig:backbone_comparison} presents a comprehensive performance comparison in terms of AUC across three datasets. Notably, the VGAE-based backbone consistently outperforms the GIN-based backbone by a substantial margin. This significant improvement can be attributed to the inherent disparities in their respective generation processes. The GIN-based backbone generates graphs deterministically, while the VGAE-based generator incorporates stochasticity.
In contrast to the GIN-based backbone, VGAE employs the reparameterization technique to learn a target distribution, allowing it to capture the data and underlying distribution. Consequently, the generated pseudo-anomalous data is more likely to reside in plausible regions, rather than simply approximating the original data. The experiment demonstrates the exceptional ability of the VGAE-based backbone to generate high-quality pseudo-anomalous data, yielding superior performance in graph-level anomaly detection tasks.
This aligns with the motivation and expectation depicted in Figure 1 of the main paper, where the stochasticity in the generation process plays an important role in learning a good decision boundary.

\end{document}